\documentclass[preprint]{elsarticle}
\usepackage{hyperref}

\journal{Expert Systems with Applications}
\date{June 1, 2021} 

\usepackage{hyphenat}
\usepackage[cmex10]{amsmath}
\usepackage{amsfonts}
\usepackage{capt-of}
\usepackage{xcolor}
\usepackage{xspace}
\usepackage{soul}
\setstcolor{red}
\usepackage[shortcuts]{extdash}
\usepackage{subcaption}

\usepackage{booktabs}
\usepackage{multirow}
\usepackage{adjustbox}

\usepackage{pifont}
\newcommand{\cmark}{\ding{51}}%
\newcommand{\ccmark}{\ding{51}{\small\kern-0.7em\ding{55}}}%
\newcommand\doublecheckmark{\checkmark\kern-0.6em\checkmark}

\biboptions{authoryear}

\begin{document}

\begin{frontmatter}
\title{Cross-Domain Object Detection Using\\Unsupervised Image Translation}

\author[ufes]{Vinicius F. Arruda\corref{mycorrespondingauthor}}
\cortext[mycorrespondingauthor]{Corresponding author}
\ead{viniciusferracoarruda@gmail.com}

\author[ufes]{Rodrigo F. Berriel}
\ead{berriel@lcad.inf.ufes.br}

\author[ufes,ifes]{Thiago M. Paix\~ao}
\ead{paixao@gmail.com}

\author[ufes]{Claudine Badue}
\ead{claudine@lcad.inf.ufes.br}

\author[ufes]{Alberto F. De Souza}
\ead{alberto@lcad.inf.ufes.br}

\author[unitn]{Nicu Sebe}
\ead{niculae.sebe@unitn.it}

\author[ufes]{Thiago Oliveira-Santos}
\ead{todsantos@inf.ufes.br}

\address[ufes]{Universidade Federal do Esp\'irito Santo (UFES), Brazil}
\address[ifes]{Instituto Federal do Esp\'irito Santo (IFES), Brazil}
\address[unitn]{University of Trento (UNITN), Italy}

\begin{abstract}
Unsupervised domain adaptation for object detection addresses the adaption of detectors trained in a source domain to work accurately in an unseen target domain. Recently, methods approaching the alignment of the intermediate features proven to be promising, achieving state-of-the-art results. However, these methods are laborious to implement and hard to interpret. Although promising, there is still room for improvements to close the performance gap toward the upper-bound (when training with the target data). In this work, we propose a method to generate an artificial dataset in the target domain to train an object detector. We employed two unsupervised image translators (CycleGAN and an AdaIN-based model) using only annotated data from the source domain and non-annotated data from the target domain. Our key contributions are the proposal of a less complex yet more effective method that also has an improved interpretability. Results on real-world scenarios for autonomous driving show significant improvements, outperforming state-of-the-art methods in most cases, further closing the gap toward the upper-bound.
\end{abstract}

\begin{keyword}
Unsupervised Domain Adaptation, Object Detection, Generative Adversarial Networks, Unpaired Image-to-Image Translation, Style-transfer
\end{keyword}
\end{frontmatter}


\section{Introduction}
\label{sec:introduction}
Object detectors based on deep learning have achieved state-of-the-art performance \citep{fasterrcnn, yolo, retinanet, efficientdet}. This outstanding performance nonetheless comes at the cost of large annotated datasets of the target application domain. In this context, an important issue is knowledge transfer from a (source) domain, for which data are easier to acquire, to a new (target) domain. This problem is known as Domain Adaptation (DA). When no annotation is available for the target domain images, the problem is referred to Unsupervised Domain Adaptation (UDA), which usually requires a semantic similarity between source and target domains. For instance, abundant data (images and annotations) may be available for day-time car detection, while night-time data are much more scarce.

In the context of self-driving vehicles, a popular research topic, objects such as crosswalks, traffic lights, pedestrians, riders, and cars, must be correctly detected in many different scenarios, such as: \textit{day-} and night-time; \textit{sunny}, rainy or snowy weather; \textit{urban}, \textit{highway} or country roads; and different camera settings (the highlighted scenarios are the most prevalent ones).

Unsupervised Domain Adaptation have been used in different types of problems, such as classification \citep{dwt, dann, coral}, segmentation \citep{task}, detection \citep{dafaster, strongweakda, arruda, scl, sapn, scda}, and more. In this work, we are interested in UDA applied to object detection. The current state-of-the-art methods of UDA for detection approach the problem very similarly to each other, by integrating domain adaptation components to an object detector (e.g., Faster R-CNN \citep{fasterrcnn}). Chen et al. \citep{dafaster} combined two domain adaptation components in the Faster R-CNN to minimize the domain shift at image- and instance-level, respectively. Claiming that such domain-shift minimization may hurt the performance for large shifts, Saito et al. \citep{strongweakda} integrated Faster R-CNN with a weak global alignment, which partially minimizes the domain-shift at image-level, and a strong local alignment. The UDA problem for object detection is far from solved. Currently, the upper-bound (when training with the target data) remains much higher than current state-of-the-art methods \citep{arruda, dafaster, strongweakda, scl, sapn, scda}, indicating a lot of room for improvement. In this work, we aim at further closing this performance gap.

In this context, we propose a simpler yet efficient two-stage approach: i) training an unsupervised image-to-image translation model to generate an artificial dataset that resembles the target domain (fake-data), and ii) training an object detector with the newly acquired data.

In a preliminary work \citep{arruda}, we showed that UDA can benefit from using an unsupervised image-to-image translation model (CycleGAN \citep{cyclegan}) to train a deep object detector (Faster R-CNN \citep{fasterrcnn}). Although the results were promising, the experiments were limited: the approach was solely evaluated in one dataset, in the day-to-night scenario, with a single object category. In addition, the preliminary work was also limited to a single image-to-image translation approach.

This work extends and consolidates \citep{arruda} by presenting an in-depth study highlighting the following points that were not considered before:

\begin{itemize}
    \item investigation of the use of a style-transfer model, in addition to CycleGAN, for unsupervised image-to-image translation to generate the fake-data;
    \item investigation of the efficacy of the method on multi-class datasets; 
    \item investigation of the benefits of using two translation models together, comprising an artificial dataset from two distinct generation processes;
    \item evaluation of the method with different arrangements of the annotated source training dataset and those artificially generated for the second-stage;
    \item conduction of an extensive comparison of the results with state-of-the-art methods, leading to a more detailed discussion;
    \item investigation of a diverse set of experiments with several well-known datasets (Cityscapes \citep{cityscapes}, KITTI \citep{kitti}, SIM10k \citep{sim10k} and Foggy Cityscapes \citep{foggycityscapes}) that includes learning from synthetic data, driving in adverse weather conditions, and cross-camera adaptation.
\end{itemize}
The experimental results have consolidated the impact of our proposed method by showing that our naive approach outperforms the sophisticated state-of-the-art methods \citep{dafaster, strongweakda} by up to $9.9$ p.p., showing the effectiveness of our simple approach.

The main contribution of our work is showing that using an image translation method in combination with a standard object detector is sufficient to surpass the accuracy of state-of-the-art methods which, in contrast, are laborious to implement. In addition, our method has as a novelty the possibility of visualizing the artificially generated data in the target domain, enabling the inspection of the quality of the target data that will be used in the training of the detector. This represents a significant advantage in terms of interpretability when compared to the state-of-the-art.

The remainder of the work is organized as follows. The next section presents the related work. Section~\ref{sec:proposed} describes the proposed method. The experimental method and the obtained results are in Section~\ref{sec:experimental}. Finally, the conclusion is presented in Section~\ref{sec:conclusion}.

\section{Related Work}
\label{sec:related_work}
This section starts with a literature review of the two main topics that give support to this work: Object Detection and Unsupervised Image-to-Image Translation. Subsequently, it presents a critical evaluation of the main problem addressed in this work: Unsupervised Domain Adaptation for Object Detection. Finally, we highlight the gap of the literature and present the advantages and limitations of our proposal.

\subsection{Object Detection}
Deep convolutional neural networks (CNNs) have proven to outperform hand-crafted filter based methods in many applications \citep{krizhevsky2012imagenet}. Hence, their usage as a feature extractor for object detection models also boosted the results \citep{fasterrcnn, yolo, retinanet}. Among the detection models, region-based CNNs excel in effectiveness. Their success gave rise to the Faster R-CNN model, which achieved consistent results, being one of the most consolidated in the literature.
Regardless of their success, neither Faster R-CNN nor other object detection models are suitable for domain adaptation tasks, requiring specific improvements to increase the accuracy. In this work we employ the Faster R-CNN without any modification as part of a two-stage approach to the UDA problem.

\subsection{Unsupervised Image-to-Image Translation}
The emergence of Generative Adversarial Networks leveraged the building of image-to-image translation models. Among them, CycleGAN \citep{cyclegan} enabled the translation of images between two distinct distributions. The approach employs two pairs of generator-discriminator, in which each generator translates the image to the opposite domain while the respective discriminator adversarially classifies the domain of the images. The model is trained to fully translate the images from a source to a target domain, completely matching the target distribution. As an advantage, this model does not require paired images during training. With a similar purpose, style-transfer techniques also adapt images to other desired domains. In \citep{adain}, the authors recently introduced an encoder-decoder approach using the proposed Adaptative Instance Normalization (AdaIN), referred from now on as AST-AdaIN. The model works by encoding a sample from both source and target domains, and mapping the mean and variance of the source feature to match those of the target domain. The adjusted feature is then decoded back to the image space, but, is resembling the style of the target domain. This approach also does not require paired images, but, unlike CycleGAN, it only tries to match low-level features of the target domain, such as textures and colors, and it may be effective only for small domain-shifts. In this work, we employed these two image translation models to generate our artificial (fake) dataset. Our choice is justified by their good results obtained without the need of paired images. Also, using them together to generate a dataset might leverage both of their advantages: while CycleGAN performs better high-level feature translations, AST-AdaIN excels in low-level feature translations.

\subsection{Unsupervised Domain Adaptation for Object Detection}
Addressing domain adaptation for object detection, Chen et al. \citep{dafaster} proposed the Domain Adaptative Faster R-CNN (DA-Faster), that tackles the domain shift at two feature levels. The proposal is built on top of the Faster R-CNN model by adding two domain adaptation components, aiming to minimize the domain discrepancy at image- and instance-level representation, respectively. Both are small networks adversarially trained to classify the domain of the feature. In short, the components enforce the domains of the features to be indistinguishable. However, as argued by Saito et al. \citep{strongweakda}, a domain-shift minimization at image-level may hurt the performance for domains with large discrepancies by also aligning background objects. Trying to address this shortcoming, Saito et al. \citep{strongweakda} proposed the Strong-Weak Distribution Alignment (Strong-Weak DA). The model employs two domain classifier networks, one at local-level, in the middle of the feature extractor, and other at image-level, right after the feature extractor. Both domain classifiers are also trained in an adversarial manner. The local-level network tries to strongly minimize the domain discrepancies only at low-level representations, such as color or texture. Conversely, the image-level network weights more on hard-to-classify examples than the easy ones, hence a weak domain alignment. However, some domains may still require a strong domain-shift minimization at image-level. In addition, the alignment of the features is not guaranteed to preserve the essential information of the target domain. Since annotations are not available at training time, the optimization procedure may find a shortcut to fool the domain classifier \citep{shortcut}.

With a similar approach to our method, Shan et al. \citep{shan2019pixel} proposed a model integrating
a domain adaptation framework strongly based on CycleGAN and an object detector very similar to Faster R-CNN. This model is trained in an end-to-end manner, leading to a large consumption of GPU resources, requiring a GPU capable to simultaneously host a CycleGAN and a Faster R-CNN during training.

Using different strategies, Shen et al. \citep{scl} proposed a gradient detach based stacked complementary losses method that uses a combination of several losses in different network layers along with gradient detach training. Zhu et al. \citep{scda} proposed a two-fold method, first by finding pertinent regions that cover the objects of interest using k-means clustering and then aligning the target regions to the source distributions in an adversarial manner. Li et al. \citep{sapn} constructed a spatial pyramid representation with multi-scale feature maps, capturing context information of objects at different scales to further be attended softly to extract features for adversarial learning. Although promising, in contrast to our method, these methods are more laborious to implement, having several additional parameters to be trained with different losses.

Despite of the efforts made by the state-of-the art methods, there is still a large gap between them and the upper-bounds (when training with target data). While aiming at reducing this performance gap, our approach has advantages and limitations. For instance, the state-of-the-art methods have the advantage of being single-stage methods, requiring the training of a single model. In contrast, our method is a two-stage approach, requiring the training of an image translation model and the generation of a novel dataset in the target domain before training the object detector. However, as a benefit, the adopted image translation models translate the image from the source to the target domain at pixel-level (raw image). With this approach, it becomes harder for the optimization procedure not to consider all the essential information. In addition, there is no need to align the features of the object detector, employed in the second stage, since the input data are already in the target domain. Also, the one-stage methods require more GPU resources as they have more parameters to train and additional losses. Conversely, our two-stage approach employs two simple models, performing one training at a time, requiring less GPU resources. Moreover, as the second stage is a simpler model (a standard Faster R-CNN), the inference is performed with higher FPS than the aforementioned methods. Although the more sophisticated methods might intuitively seem to perform better than the simpler two-stage methods (like the one proposed), our experiments show that this is not true for most of the evaluated scenarios.

\section{Proposed Method}
\label{sec:proposed}
\begin{figure}[ht]
	\centering
	\includegraphics[width=\textwidth]{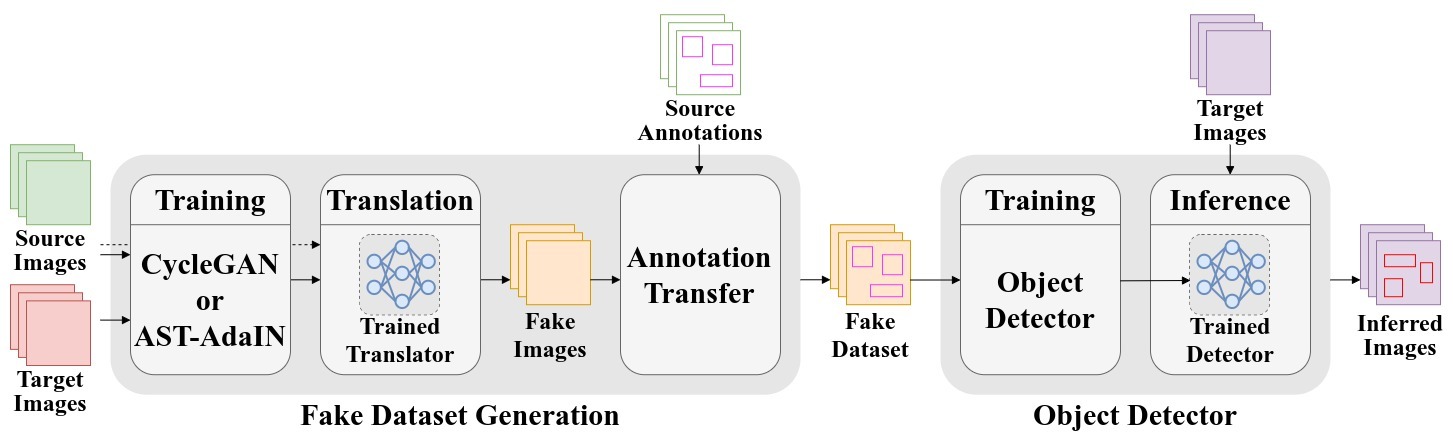}
	\caption{Overview of the proposed method. Firstly, an unsupervised image translation model is trained with unpaired source and target images. Then, the source image set is translated to its fake-target version. The annotations of source images are directly transferred to the fake-target images, composing the fake-target dataset. Finally, an object detector is trained resulting in an object detector trained on a domain without previous annotations.}
    \label{method-overview}
\end{figure}
The proposed method, illustrated in \autoref{method-overview}, is two-fold. Firstly, an image translation model is trained with unpaired images of source and target domains, to artificially generate images in the target domain. The objective of the model is to translate only the appearance between the domains, retaining unaltered the location, pose and co-occurrences of the objects. Therefore, the fake-images annotations (classes and bounding boxes) can be inherited from their respective versions in the source domain, comprising together the fake-target dataset. The second step is training an object detector with the artificially-generated dataset to detect objects in a domain which an annotated dataset was not previously available. Finally, the trained object detector can be used to infer objects in the real images of the target domain.

As the image translation and the detection are decoupled procedures, the computational performance of the inference (time spent to infer each image) depends only on the chosen object detection model, Faster R-CNN in this work.

\subsection{Fake Dataset Generation}
The fake dataset generation produces fake-images which resemble the target domain along with their respective annotations, providing training data for the object detector. This approach requires a set of annotated real images from the source domain and non-annotated real images from the target domain.
The process is given in two steps: firstly, a model for unpaired image translation is trained with the annotated source images and non-annotated target images. Secondly, the trained model is used to produce the fake-target images by translating the source images used in the training step. Their respective annotations are automatically inherited from the source domain. It is worthy to mention that since our goal is to obtain a fake-target dataset, we do not require that the translation model generalizes to other unseen images during the training phase. Also, once we acquired the fake-target dataset, the translation model is no longer necessary and can be discarded. In this work, we employed two models of image translation: an unpaired image-to-image translator and a style-transfer model.

\subsubsection{CycleGAN}
\textbf{Training.}
For the unpaired image-to-image translation, we use the CycleGAN model. The CycleGAN is trained without supervision by processing two unpaired sets of images from the source and target domains.
The model comprises two siblings generators, $G_T$ and $G_S$, where $G_T$ translates images from the source to the target domain and $G_S$ translates back from the target to the source domain completing the cycle. Simultaneously, two siblings discriminators, $D_T$ and $D_S$, are trained to classify whether a given image is coming from a real or a generated set, where $D_T$ is responsible to distinguish between the real target images and those produced by $G_T$. Likewise, $D_S$ and $G_S$ are trained in the same manner but in the source domain. $G_T$ and $G_S$ are trained to produce confident images aiming to fool $D_T$ and $D_S$, respectively. To encourage the coherence of the generated images, a cycle-consistency constraint \citep{cyclegan} is added to the loss in order to enforce to recover the source/target image when is translated back from the target/source domain. The cycle-consistency loss is defined as $|G_S(G_T(s)) - s|$ and $|G_T(G_S(t)) - t|$, where $s$ and $t$ are real source and target image samples from the training set, respectively.

\textbf{Translation.}
With the CycleGAN trained, the generator $G_T$ is used to translate each source image used in the training set to a corresponding fake-target image. Assuming the coherence between the translated image and its original, the annotations of the corresponding original images can be directly reused by the fake-image. Finally, the fake-target images and their respectively inherited annotations comprise the fake-target dataset.

\subsubsection{AST-AdaIN}
\textbf{Training.}
We used AST-AdaIN for the style-transfer. As CycleGAN, the AST-AdaIN model is also trained with unpaired images from both source and target domains. An encoder $E(\cdot)$ receives the unpaired images from source and target domains, generating their respective features maps: $E(s) = h_s$ and $E(t) = h_t$, where $s$ and $t$ are the source and target images, respectively, with their respective representations in the feature space $h_s$ and $h_t$. The Adaptative Instance Normalization layer (AdaIN) aligns the mean and variance of the source feature maps to those of the target feature maps, producing the fake-target feature maps $\hat{h}_t$:
\begin{equation}
    \hat{h}_t = AdaIN(h_s, h_t) = \sigma(h_t)\left(\frac{h_s - \mu(h_s)}{\sigma(h_s)}\right) + \mu(h_t)
    \label{eq:ht}
\end{equation}

A decoder $D(\cdot)$ is trained to decode features from the feature space to the image space, thus enabling the generation of a fake-target image, $D(\hat{h}_t)$. The training of the encoder-decoder relies on a content and a style losses. The content loss aims to minimize the difference between the aligned feature maps with the encoded feature maps of the fake-target feature maps, as following:
\begin{equation}
    \mathcal{L}_c = \left\lVert E(D(\hat{h}_t)) - \hat{h}_t\right\rVert_2
\end{equation}
The style loss aims to minimize the difference between the mean and variance features along the encoder layers, as following:
\begin{equation}
    \mathcal{L}_s = \sum_{i=1}^{L} \left\lVert \mu(\phi_i(D(\hat{h}_t))) - \mu(\phi_i(t))\right\rVert_2 + \sum_{i=1}^{L} \left\lVert \sigma(\phi_i(D(\hat{h}_t))) - \sigma(\phi_i(t))\right\rVert_2
\end{equation}
where each $\phi_i$ denotes a layer of the network used to compute the loss. 

\textbf{Translation.}
After the training of the AST-AdaIN, the model is ready to produce the fake-target images. 
Each image from the initial training dataset belonging to the source domain is fed into $E(\cdot)$, producing the $h_s$. To produce the $h_t$, we input into $E(\cdot)$ a randomly chosen image from the target domain dataset. Then, the AdaIN is employed and the resulting features are decoded, $D(AdaIN(h_s, h_t))$, producing a new and corresponding fake-target image.

In practice, a style weight can be used to control how much of the style should be transferred:
\begin{equation}
    output = D((1 - \alpha)h_s + \alpha AdaIN(h_s, h_t))
    \label{eq:output_adain}
\end{equation}
where the input image is recovered when $\alpha = 0$ and the most stylized image is synthesized when $\alpha = 1$. Likewise CycleGAN, image coherence between the source and target images was assumed,
thus the fake-target images can inherit their corresponding source images annotations, comprising the fake-target dataset.

\subsection{Object Detector}
\textbf{Training.}
The detection stage uses a general purpose object detector to locate objects in the images. The training is performed with the previously generated fake-target dataset comprising a set of images and their respective annotations (bounding box along with its class). Among several existing models \citep{fast, fasterrcnn, yolo, retinanet}, this work adopts Faster R-CNN, which besides being one of the state-of-the-art deep detectors, enables fair comparison with \citep{dafaster} and \citep{strongweakda}.

\textbf{Inference.}
Finally, with the trained object detection model, scenes in the target domain can be addressed. For each real target image, the detector predicts the bounding boxes of the objects and their respective class with a confidence score.

\section{Experimental Methodology}
\label{sec:experimental}    
The general purpose of the experiments is to evaluate (and compare with the state-of-the-art) the performance and robustness of our method in three challenging domain-shift scenarios proposed in previous works \citep{dafaster, strongweakda}: (i) learning from synthetic data, in which annotation data are only available for synthetic rendered images (source domain) and the model has to detect objects on real-world images (target domain); (ii) driving in adverse weather conditions, in which the source domain comprises sunny day images and the target domain comprises foggy images; and, finally, (iii) cross-camera adaptation, in which the images of the source and target domains have different aspect ratio, contrast, white balance, field-of-view, etc.. Since our focus is on autonomous driving, we used popular traffic-related datasets for evaluation of computer vision methods for self-driving cars. The following sections detail, respectively, the datasets, the performance metric, the setup for training the models, and the computational platform on which the experiments were carried out. Finally, we describe the experiments and present the results alongside with a discussion.

\subsection{Datasets}
\label{sec:datasets}
\begin{figure}[ht]
	\centering
	\begin{subfigure}{\linewidth}%
	    \centering%
	    \includegraphics[width=0.325\linewidth]{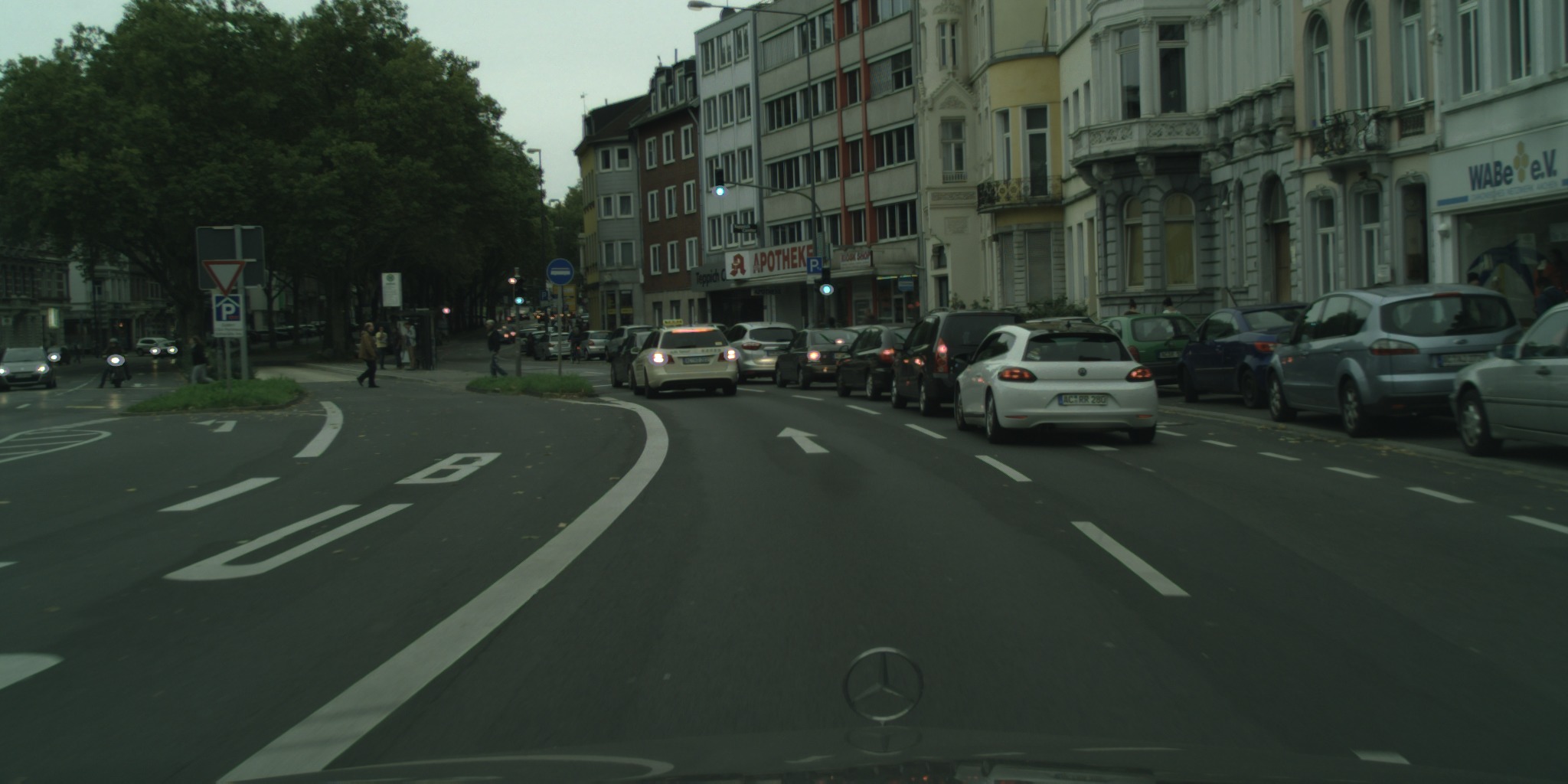}%
	    \hspace{1.5pt}%
	    \includegraphics[width=0.325\linewidth]{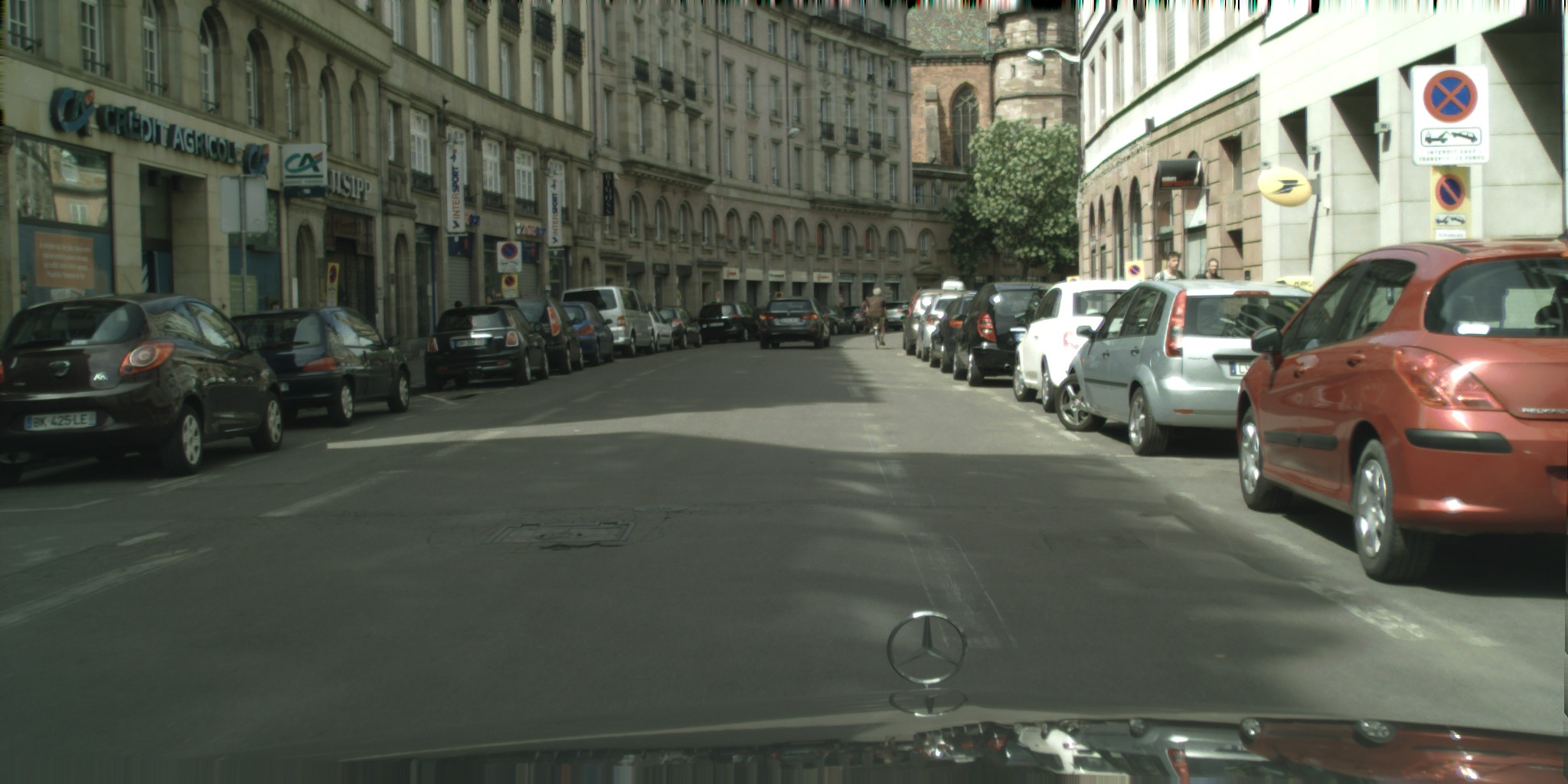}%
	    \hspace{1.5pt}%
	    \includegraphics[width=0.325\linewidth]{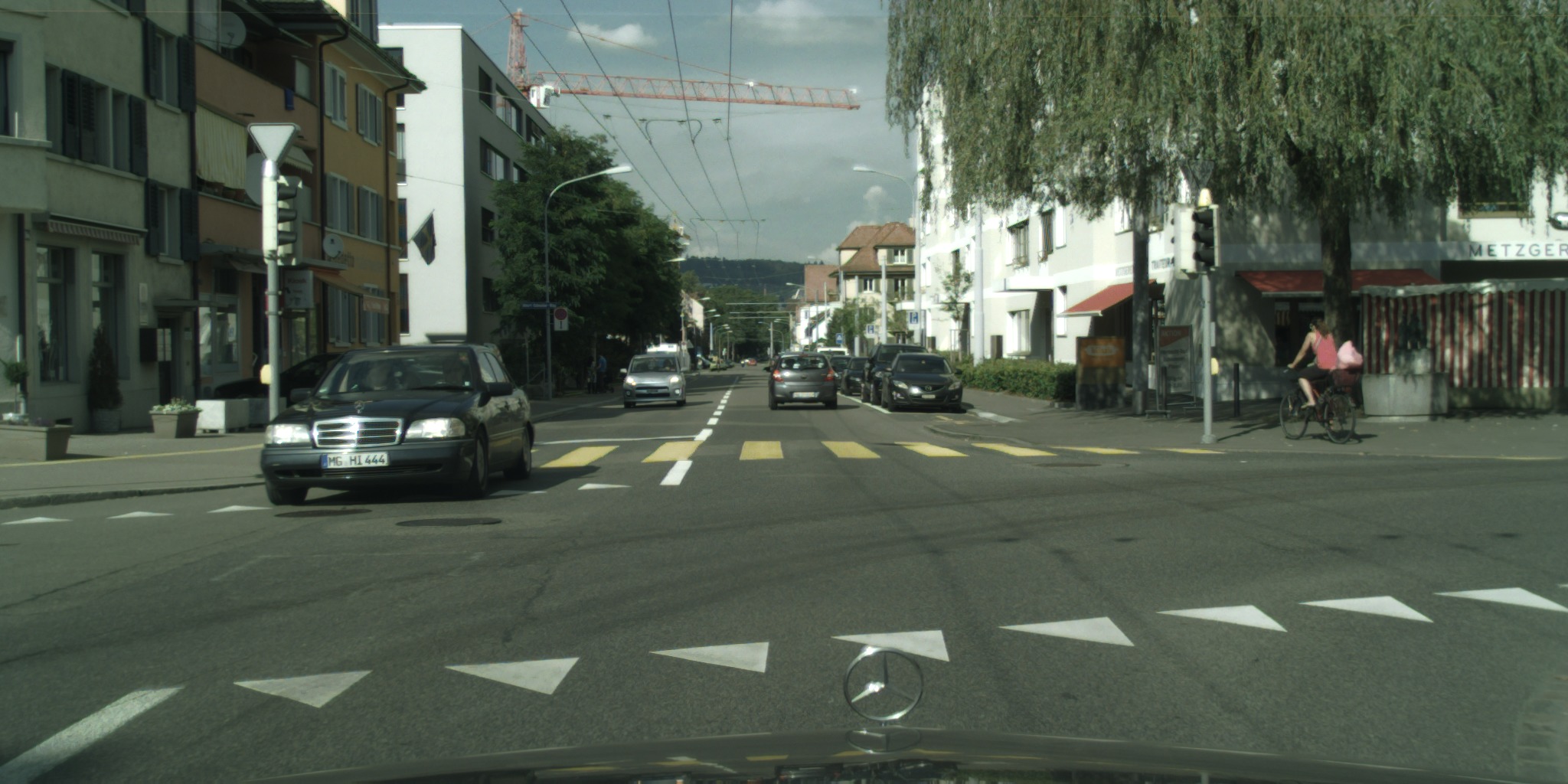}%
	\end{subfigure}%
	\\
	\vspace{2pt}%
	\begin{subfigure}{\linewidth}%
	    \centering%
	    \includegraphics[width=0.325\linewidth]{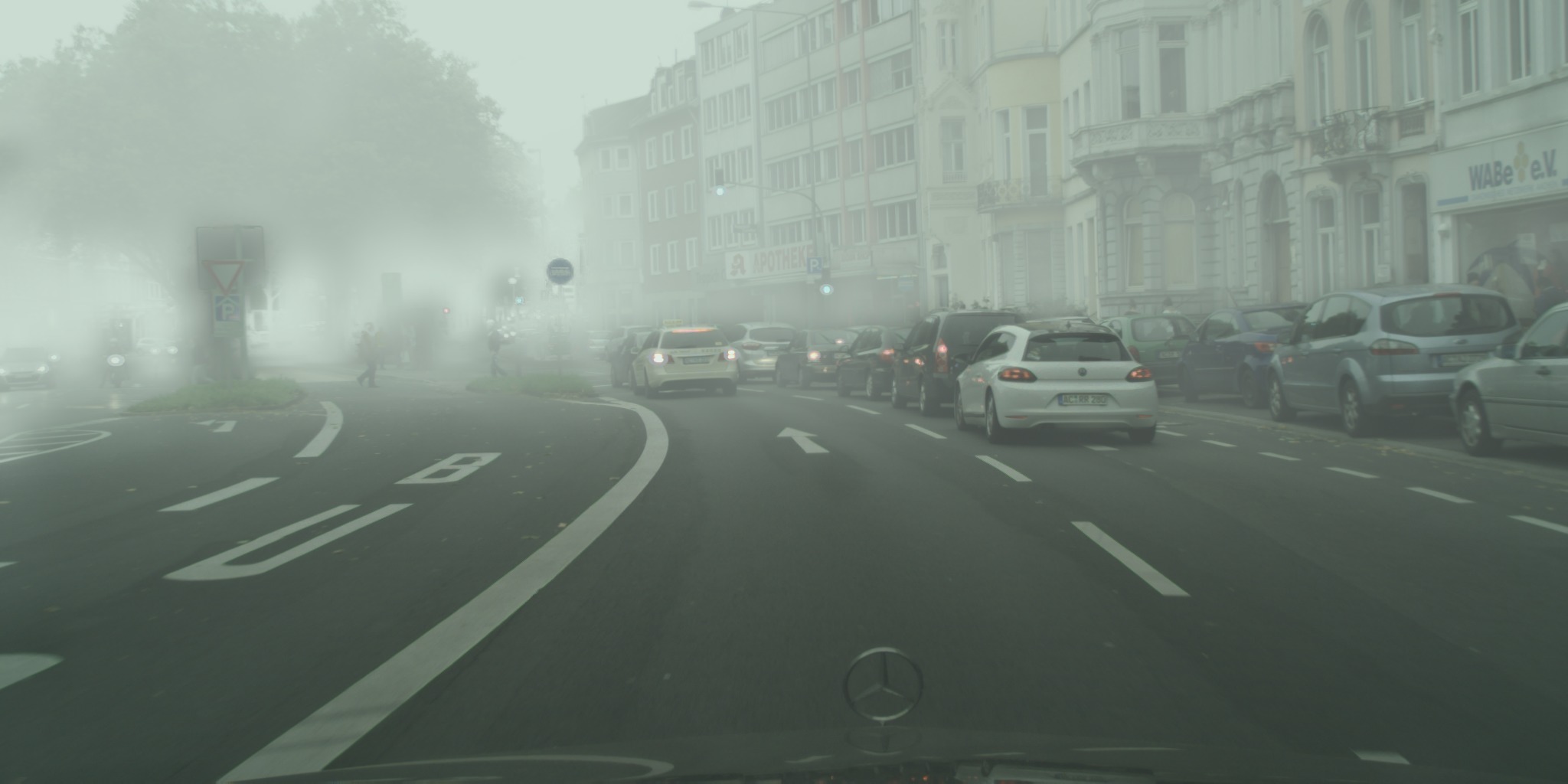}%
	    \hspace{1.5pt}%
	    \includegraphics[width=0.325\linewidth]{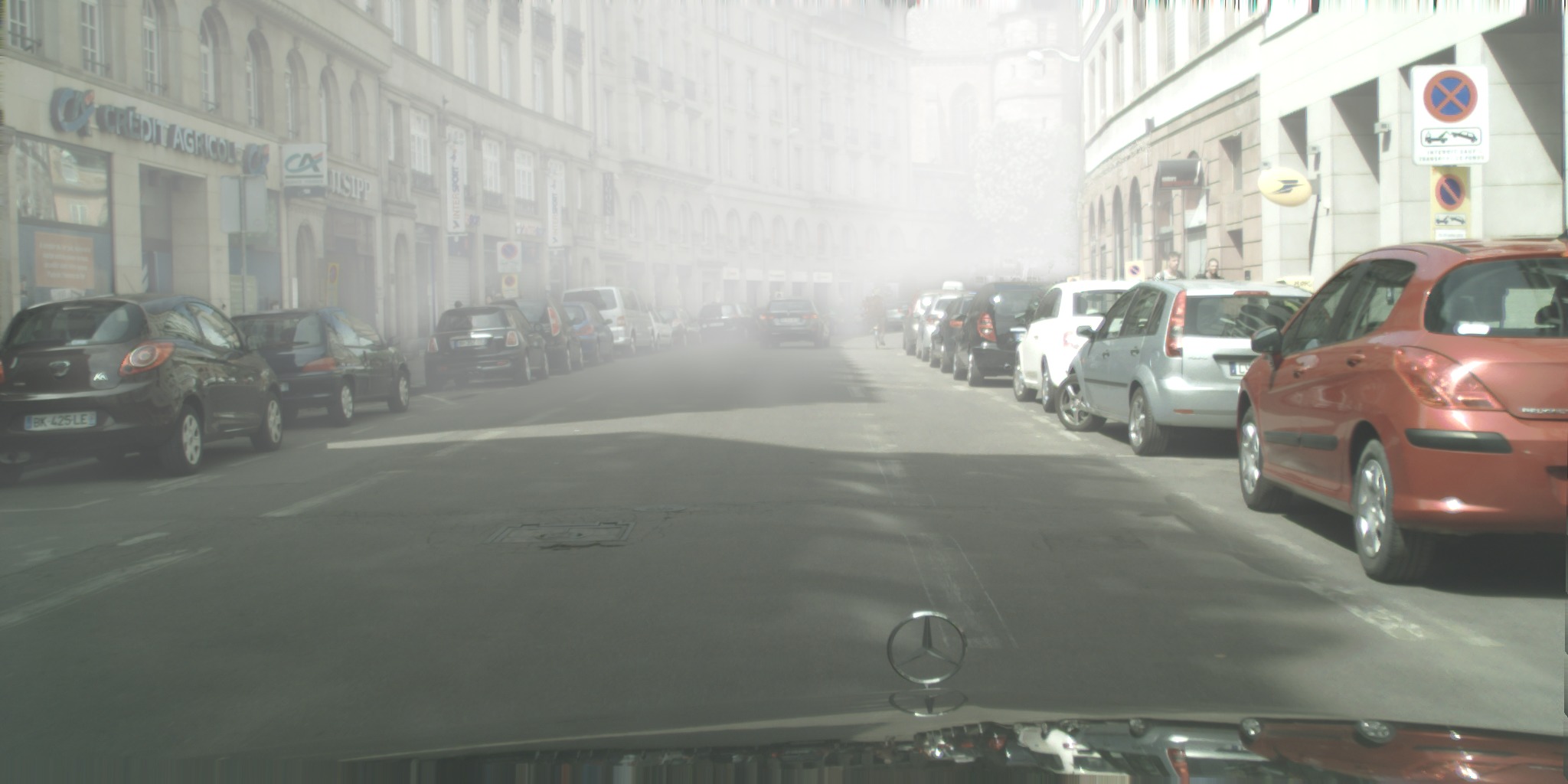}%
	    \hspace{1.5pt}%
	    \includegraphics[width=0.325\linewidth]{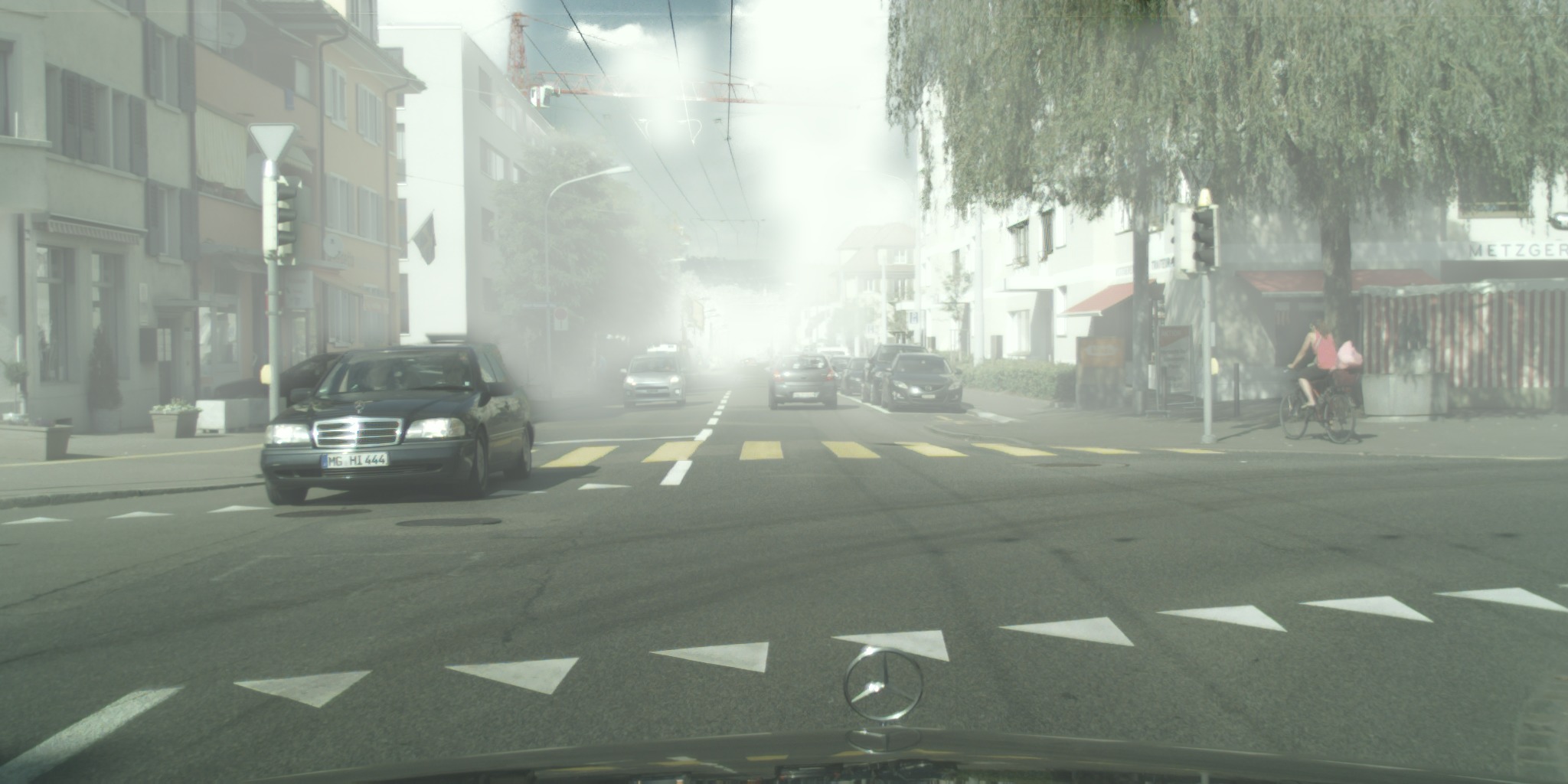}%
	\end{subfigure}%
	\\
	\vspace{2pt}%
	\begin{subfigure}{\linewidth}%
	    \centering%
	    \includegraphics[width=0.325\linewidth]{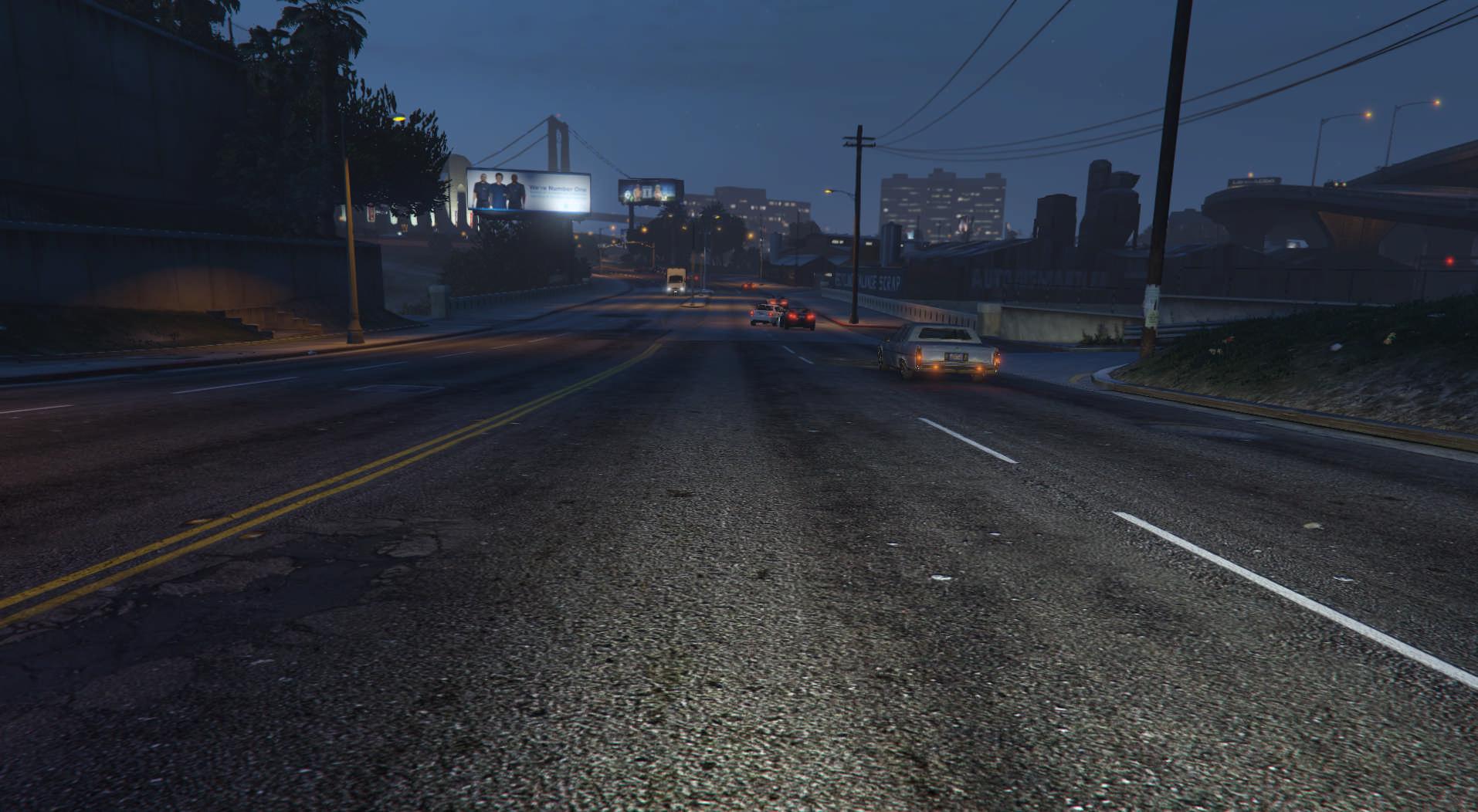}%
	    \hspace{1.5pt}%
	    \includegraphics[width=0.325\linewidth]{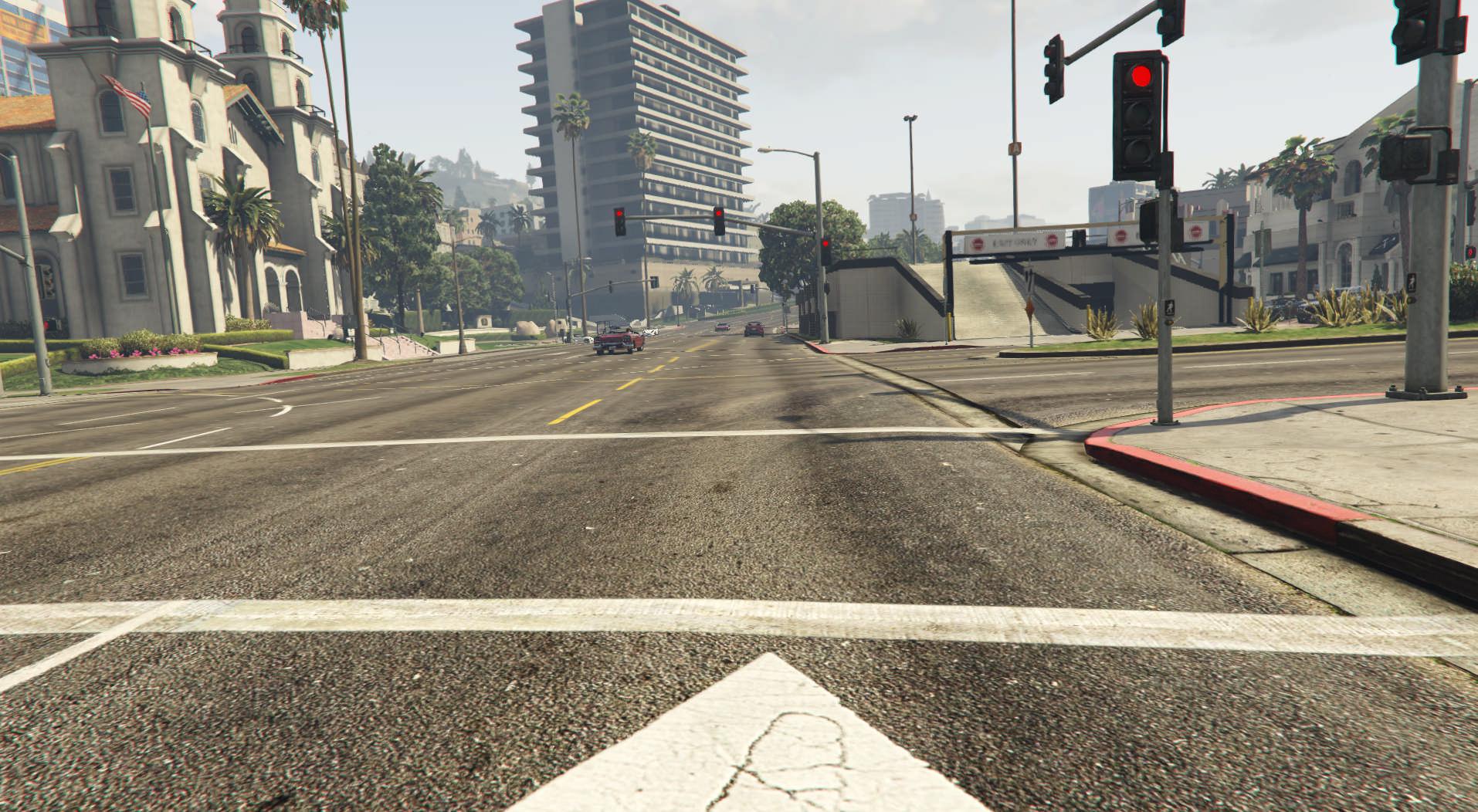}%
	    \hspace{1.5pt}%
	    \includegraphics[width=0.325\linewidth]{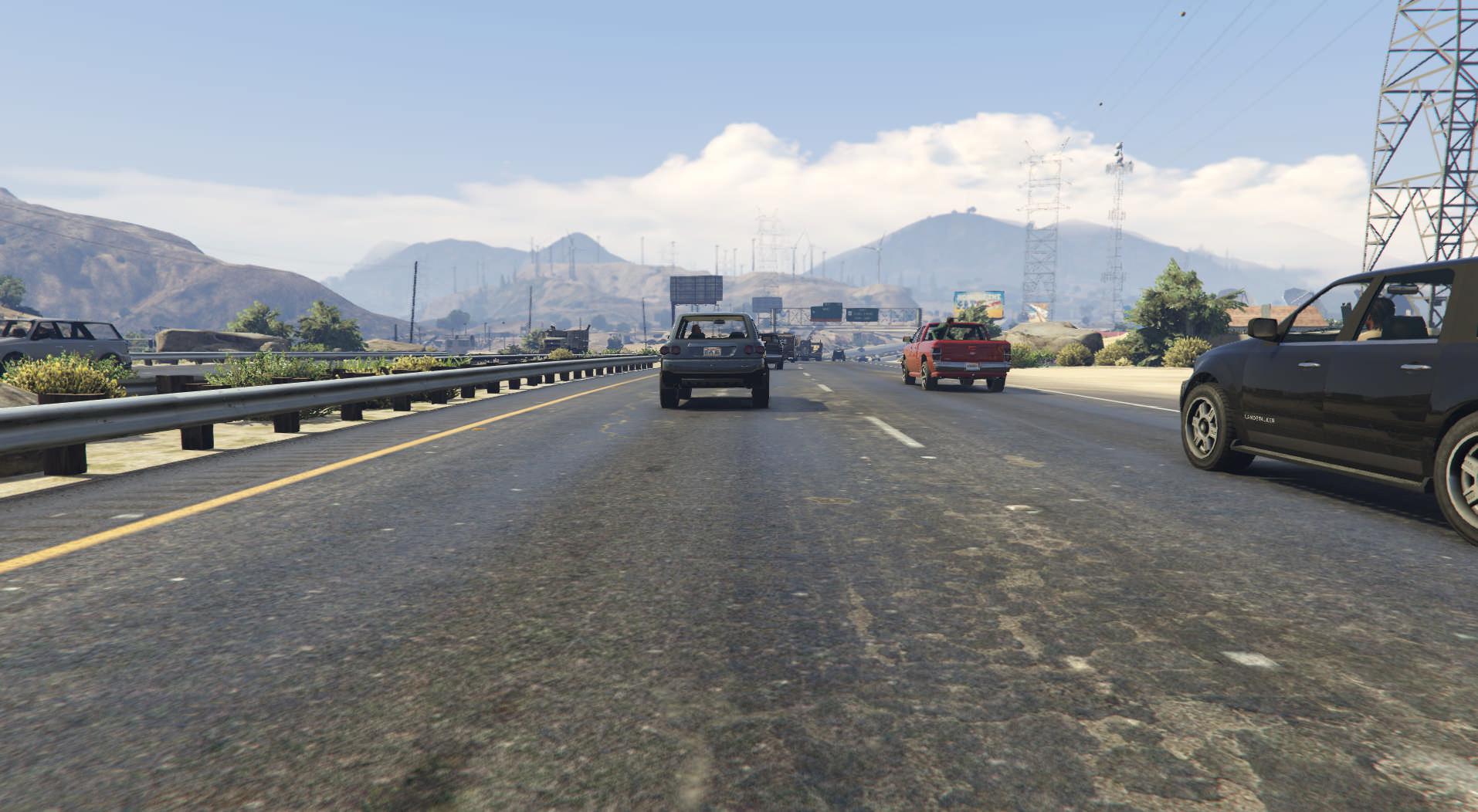}%
	\end{subfigure}%
	\\
	\vspace{2pt}%
	\begin{subfigure}{\linewidth}%
	    \centering%
	    \includegraphics[width=0.325\linewidth]{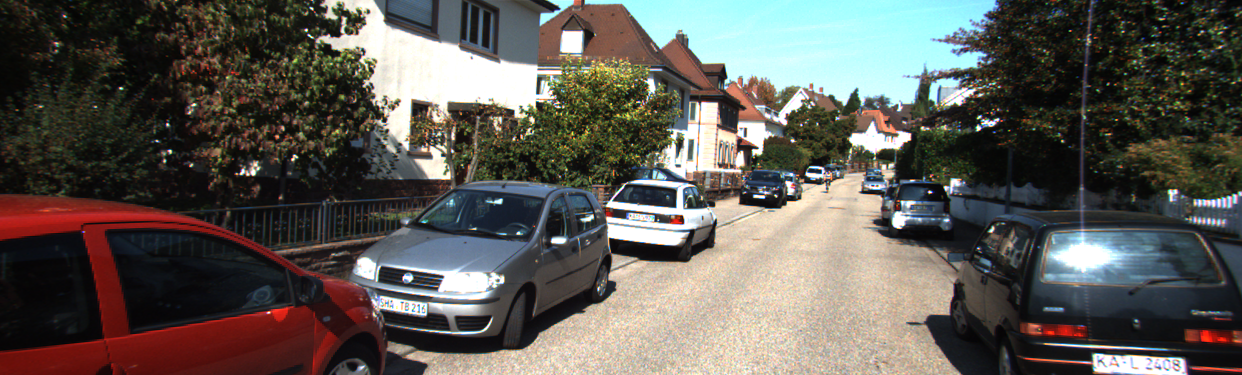}%
	    \hspace{1.5pt}%
	    \includegraphics[width=0.325\linewidth]{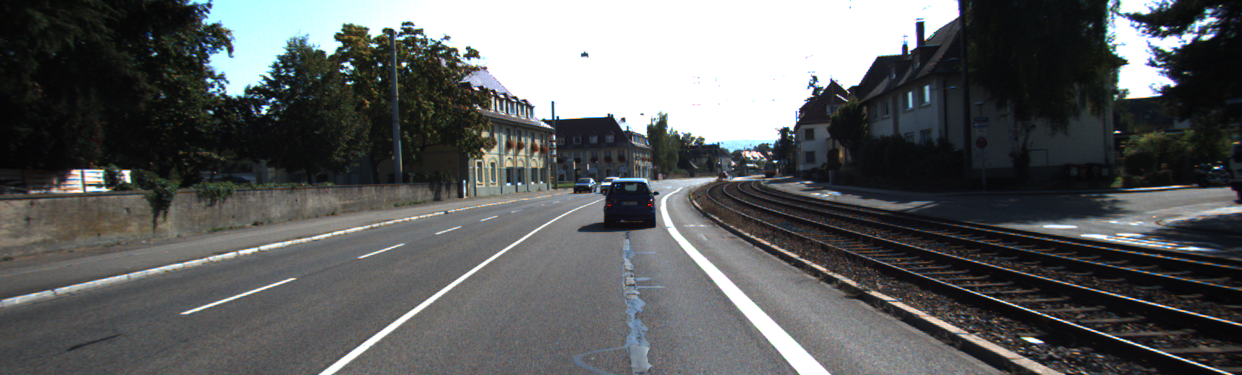}%
	    \hspace{1.5pt}%
	    \includegraphics[width=0.325\linewidth]{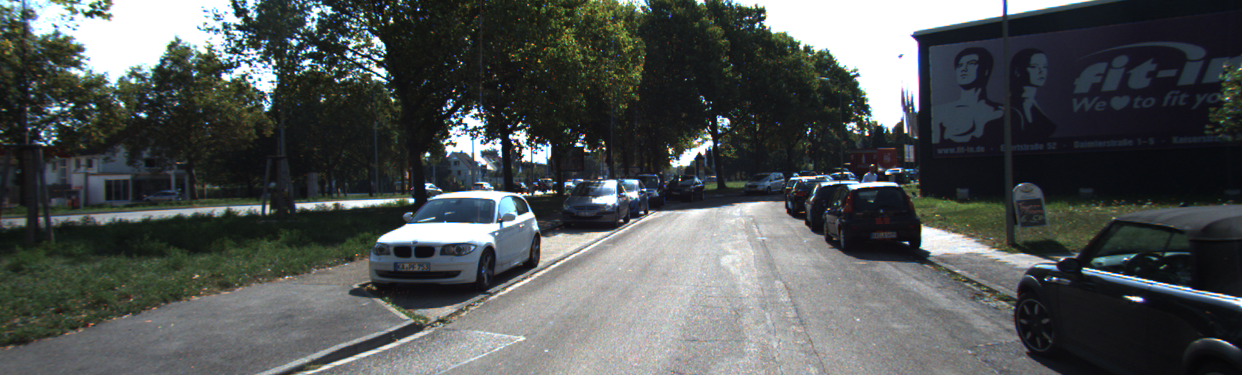}%
	\end{subfigure}%
	\caption{Samples of each dataset. The datasets samples are presented row-by-row in the following order (top to bottom): Cityscapes, Foggy Cityscapes, Sim10k and KITTI. The original aspect ratio of the images was preserved.}
	\label{fig:dataset_samples}
\end{figure}

Three datasets were used in the experiments. For the first scenario (learning from synthetic data), synthetic rendered images come from the Sim10k dataset \citep{sim10k} and real images come from Cityscapes \citep{cityscapes} (Sim10k $\rightarrow$ Cityscapes). For the second scenario (driving in adverse weather), source and target images are from Cityscapes and Foggy Cityscapes \citep{foggycityscapes}, respectively (Cityscape $\rightarrow$ Foggy Cityscape). Cross-camera adaptation includes two subscenarios: (i) Cityscapes' images as source and KITTI's \citep{kitti} as target (Cityscapes $\rightarrow$ KITTI) and (ii) vice-versa (KITTI $\rightarrow$
Cityscapes). Details of the datasets are provided in the following subsections.

\subsubsection{Cityscapes}
The Cityscapes \citep{cityscapes} dataset comprises urban 2048$\times$1024 images for driving scenarios. The images were acquired using a car-mounted video camera while driving through several European cities with diverse scenarios. The original annotations are instance segmentation of the objects, thus it was necessary to adapt the annotations to bounding boxes coordinates as in \citep{dafaster, strongweakda} by taking the tightest rectangles of the instance masks. The dataset is originally divided in 2,975 images for training and  500 for validation. For our purposes, the training partition is used to train the adaptation models and to train the Faster R-CNN, whereas the validation images are used to test the detector. The dataset comprises a total of 8 object categories: person, rider, car, truck, bus, train, motorcycle and bicycle. The top row of \autoref{fig:dataset_samples} shows some samples from the training set.

\subsubsection{Foggy Cityscapes}
Foggy Cityscapes \citep{foggycityscapes} is an artificial dataset rendered with computer graphics techniques to add fog in the Cityscapes' images. The annotations, object classes and dataset split are the same as in the original Cityscapes dataset. To illustrate, the second row of \autoref{fig:dataset_samples} shows the same samples from the training set of the Cityscapes in the foggy version.

\subsubsection{Sim10k}
Sim10k \citep{sim10k} is a synthetic dataset rendered by the game Grand Theft Auto V.  The annotations are only available in the training partition. The dataset comprises 10,000 1914$\times$1052 images with 58,701 bounding-box annotations of cars. Samples are shown in the third row of \autoref{fig:dataset_samples}.

\subsubsection{KITTI}
KITTI \citep{kitti} comprises 1250$\times$375 images from several traffic objects, and it is originally divided into training and test sets, and only the training set is annotated. Since annotations are required for both training and evaluating the models, only the training images (7,481 samples) were used. Following previous work \citep{dafaster}, only annotations regarding cars were considered. Samples are shown in the last row of \autoref{fig:dataset_samples}.

\subsection{Performance Metric}
\label{sec:metrics}
The performance of the object detector is measured by the mean Average Precision (mAP). Clearly, mAP has become the most popular measure for evaluating object detectors in the academic literature \citep{fast, fasterrcnn, yolo, efficientdet}, as well as in the main related competitions: PASCAL Visual Object Classes Challenge (PASCAL VOC) \citep{pascal2007, pascal2012, pascal2015} and Common Objects in Context Challenge (COCO) \citep{coco}. The mAP summarizes the precision and recall curves into a single value, therefore taking into consideration both properties simultaneously. Unlike the traditional measures such as precision, recall, and F1 measures, mAP has the advantage of being independent of a particular threshold for the confidence score, which enables a fairer comparison.

The Average Precision (AP) of each object class is defined as the area under the precision-recall curve. This curve is built by calculating the precision and recall values of the accumulated true positive or false positive detections. For this, detections are ordered by their confidence scores, and precision and recall are calculated for each accumulated detection. Given the precision-recall curve, for the 2007 version (VOC'07) \citep{pascal2007}, interpolated precision values are measured for 11 evenly divided recall levels, while for the 2012 version (VOC'12) \citep{pascal2012}, interpolated precision values are measured for all recall levels. For this, for each recall level $r$, it is taken the maximum precision whose recall value is greater or equal than $r + 1$. Then, AP is calculated as the total area under the interpolated precision-recall curve. Finally, the mAP
is calculated as the mean of the AP of all classes. As mAP is the mean of the APs across all classes, if only one class is provided, the mAP will be the same as the AP of the class.

While the works of the state-of-the-art methods \citep{dafaster, strongweakda} use the VOC'07 definition of the mAP, in this work we used the definition proposed in the PASCAL VOC Challenge 2012 (VOC'12) \citep{pascal2012}, because it improves the precision and ability to measure differences between methods with low AP. 

To keep the results comparable, we also computed the results of the state-of-the-art methods with the VOC'12 metric. It is important to note that this update of the metric does not worsen the comparison, but makes it more accurate.

\subsection{Training Setup}
In this section, the details and hyperparameters used for training the translation and detection models are presented.

\subsubsection{Faster R-CNN}
\label{subsec:baselines}
The Faster R-CNN feature extractor was initialized with ResNet-101 \citep{resnet} pre-trained on the ImageNet dataset \citep{imagenet}. Anchor scales and ratios were defined considering the application working range as $\{4, 8, 16, 32\}$ and $\{0.5, 1, 2\}$, respectively. For experiments with only one dataset, the learning rate was kept fixed to $0.001$ for the first 50k iterations and linearly decayed to zero over the next 20k iterations, resulting in 70k iterations. For the experiments with two or more datasets, the same learning rate was kept fixed for the first 70k iterations and linearly decayed to zero over the next 30k iterations, resulting in 100k iterations. The training was carried out with one image per batch, each one having its smaller side resized to $500$ pixels keeping the aspect ratio. Also, data-augmentation was performed by flipping the images horizontally with probability of $50\%$. A public source code\footnote{\url{https://github.com/endernewton/tf-faster-rcnn}} was used for carrying out the experiments.

\subsubsection{CycleGAN}
The architecture used was the same as in the original paper \citep{cyclegan}, except by the 16$\times$16 PatchGAN \citep{cyclegan, pix2pix} discriminator that has a smaller field of view (and fewer layers) than the original 70$\times$70 PatchGAN discriminator. The model was trained with 100 epochs using a fixed learning rate of $0.0002$ followed by 100 epochs with learning rate linearly decaying to zero, totaling 200 epochs. Only one image was used per batch, each one loaded and resized to 572$\times$572 pixels (the aspect ratio may be deformed) followed by a random crop of 512$\times$512 pixels. Data augmentation was employed by flipping the images horizontally with probability of $50\%$. The default values were used on the other hyperparameters. The adopted source code was developed by the CycleGAN's authors, and is publicly available\footnote{\url{https://github.com/junyanz/pytorch-CycleGAN-and-pix2pix}}.

\subsubsection{AST-AdaIN}
For the Arbitrary Style Transfer with AdaIN, we used the same architecture as in the original paper \citep{adain}. During training, the input images are resized to 512$\times$512 pixels followed by random cropping to 256$\times$256 pixels. At test time, the images' dimensions are not modified since the model is fully convolutional. The training is conducted for 160k iterations and the images are processed in batches of 8 images. All other hyperparameters were set to the default values. For transferring the style from a source to a target image, we adopted $\alpha = 1$ (\autoref{eq:output_adain}) to generate fully stylized images. The source code employed in the experimentation is publicly available\footnote{\url{https://github.com/naoto0804/pytorch-AdaIN}}.

\subsubsection{State-of-the-art Models}
For comparison purpose, the state-of-the-art models DA-Faster \citep{dafaster} and Strong-Weak-DA \citep{strongweakda} were used, training with the same hyperparameters described in their respective works. Among the four datasets experimented in this work, all four were adopted in the DA-Faster work and three in the Strong-Weak-DA work. Considering that all experiments conducted in their work employed the same set of hyperparameters, it was decided to keep their proposed values with the assumption that these were the best hyperparameters found by the authors. As our method uses ResNet-101 pre-trained on the ImageNet dataset, the backbone of the state-of-the-art models were changed from pre-trained VGG-16 to ResNet-101. The code used is publicly available for both models\footnote{DA-Faster: \url{https://github.com/tiancity-NJU/da-faster-rcnn-PyTorch}}\textsuperscript{,}\footnote{Strong-Weak-DA: \url{https://github.com/VisionLearningGroup/DA_Detection}}.

\subsection{Experimental Platform}
The experiments were carried out with an Intel Xeon E5606 2.13 GHz x 8 with 32 GB of RAM, and 1 Titan Xp GPU with 12 GB of memory. The machine was running Linux Ubuntu 16.04 with NVIDIA CUDA 9.0 and cuDNN 7.0~\citep{cudnn} installed. The training and inference steps were done using the TensorFlow~\citep{tensorflow} and PyTorch~\citep{pytorch} frameworks, depending on the model. The training sessions took, on average, 3 weeks for the CycleGAN model, 20 hours for the AST-AdaIN model, and 12 hours for the Faster R-CNN model. For the state-of-the-art models (DA-Faster and Strong-Weak-DA), 24 training hours were taken, on average. In the used setup, CycleGAN translates images at an approximate rate of 6 frames-per-second (fps), AST-AdaIN transfers the style of the images at a rate of 11 fps, whereas Faster R-CNN performs detections at $\approx 7$ fps. For comparison, due to the extra parameters, the state-of-the-art methods run at approximately 5 fps. In contrast, our simpler approach performs detections $1.4 \times$ faster. Training scripts, pre-trained models, and dataset will be made publicly available upon acceptance.

\subsection{Experiments and Results}
The evaluation of the object detectors was carried out on the test partition of the target domain dataset. For some tasks, the detection system might need to work well in both source and target domains, e.g., sunny and foggy driving scenarios in the same day, therefore we also evaluated the test sets of both domains when applicable.

For our approach, different training settings were evaluated according to the data available. These settings combine fake data, generated by the image translation process, and the source-domain training data, assembling up to three datasets into each setting. The training settings receive an acronym in the format: OURS-\{S, C, A\}, and are described in \autoref{tab:acronym}.

\begin{table}[t]
    \centering
    \begin{tabular}{c | c | c c} 
     \toprule
     Training & Real & \multicolumn{2}{c}{Fake} \\
     Setting & \underline{S}ource & \underline{C}ycleGAN & \underline{A}ST-AdaIN \\
     \midrule\midrule
     OURS-C & & \cmark & \\
     OURS-A & & & \cmark \\
     OURS-S+C & \cmark & \cmark & \\
     OURS-S+A & \cmark & & \cmark \\
     OURS-C+A & & \cmark & \cmark \\
     OURS-S+C+A & \cmark & \cmark & \cmark \\
     \bottomrule
    \end{tabular}
    \caption{
    Description of the training settings used in the experiments. The training settings are assembled from the annotated data that are available: real source training set (S), CycleGAN fake-data (C), and AST-AdaIN fake-data (A), comprising up to six different training settings.}
    \label{tab:acronym}
\end{table}

For comparison, we trained: the (i) source- and (ii) target-only as an empirical measure of the lower- and upper-bound, respectively, and the state-of-the-art models (iii) DA-Faster, and (iv) Strong-Weak DA. For source- and target-only, Faster R-CNN was trained directly with the source and target training sets, respectively, following the setup described in Section \ref{subsec:baselines}. The source-only was trained only in the source domain training set. The target-only, when evaluating only in the target domain, was trained only in the target domain, and when evaluating simultaneously on both test sets of source and target domains, the training was performed on the source and target domains training sets altogether.

\subsubsection{Learning from synthetic data}
This experiment intends to measure the ability of our method in adapting to real-world data while having access only to the annotations of the synthetic data. For this, the source and target domains datasets were Sim10k and Cityscapes, respectively. Following previous works \citep{dafaster, strongweakda}, we considered only the car class in this experiment as it is the only class present in both datasets simultaneously. All the available training data was used (10k annotated images from Sim10k and 2,975 non-annotated images from Cityscapes) for training the adaptation models.
The evaluation of the detector was only performed on the target domain since the system is not expected to run on synthetic scenarios like Sim10k.

In this work, due to the unavailability of the source-code, it was not possible to directly compare with the method proposed by Shan et al. \citep{shan2019pixel} and Zhu et al. \citep{scda}. Also, other works have recently appeared (\citep{scl, sapn}) but, despite the availability of the source-code, the requirement of editing them to match our settings may lead to a biased result. Although the experiments reported here are also present in their work, their method was built with the VGG-16 as the backbone and the results were reported using the VOC'07 metric. Exceptionally in this section, we added the results experimenting with our method matching their settings, using the VGG-16 as the backbone, and evaluating with the VOC'07 metric.

\textbf{Results.}
\begin{table}[t]
    \centering
    \begin{tabular}{c | c c | c c | c} 
     \toprule
      & \multicolumn{2}{c}{Real} & \multicolumn{2}{|c|}{Fake} & \\
     Method & S & T & C & A & car AP \\
     \midrule\midrule
     Source-only & \cmark & & & & 34.7\\ 
     \hline
     DA-Faster \citep{dafaster} & \cmark & \ccmark & & & 27.3 \\
     Strong-Weak DA \citep{strongweakda} & \cmark & \ccmark & & & 31.7 \\
     \hline
     OURS-C & & & \cmark & & 42.4 \\
     OURS-A & & & & \cmark & 43.7 \\
     OURS-C+A & & & \cmark & \cmark & \textbf{44.6}\\
     \hline
     Target-only & & \cmark & & & 58.9 \\
     \bottomrule
    \end{tabular}
    \caption{
    Sim10k$\protect\rightarrow$Cityscapes:
    Results from training with annotated Sim10k (S) and non-annotated Cityscapes (T), evaluating on Cityscapes.
    The check-mark denotes which data was used during training, and if crossed, only non-annotated data was used.
    Fake data was acquired by translating Sim10k to Cityscapes using CycleGAN (C) and AST-AdaIN (A).}
    \label{tab:results-synthetic}
\end{table}
The detection results are shown in \autoref{tab:results-synthetic}. As it can be seen, there is a clear performance gap of $24.2$ p.p. ($58.9$ vs. $34.7$) between the empirical lower- and upper-bounds, indicating a need for methods that handle this domain adaptation problem. Our method outperformed source-only and both state-of-the-art models. Although training with the fake-data generated by both translation methods (OURS-C+A) attained the best result with $44.6\%$ in AP (an increment of $9.9$ p.p. over the highest baseline result), the simpler training settings (OURS-C and OURS-A) also surpassed the other models (for results using also the source domain to train, please see our supplementary material).

The AST-AdaIN method achieved a higher AP ($+1.3$ p.p.) when compared to the CycleGAN ($43.7$ vs. $42.4$), which shows that it better translates synthetic to real images. This result is in line with the intuition that this scenario requires a more accurate alignment at the low-level image features, such as color and texture. Although Strong-Weak DA promises strong low-level alignment, it may not preserve the essential (semantic) information. This suggests that pixel-level translation methods should be preferred over Strong-Weak DA. The lower performance of the DA-Faster model w.r.t. the source-only has already been reported in \citep{strongweakda}.

\begin{figure}[t!]
	\centering
	\begin{subfigure}{0.75\linewidth}
	    \centering
	    \includegraphics[width=\textwidth]{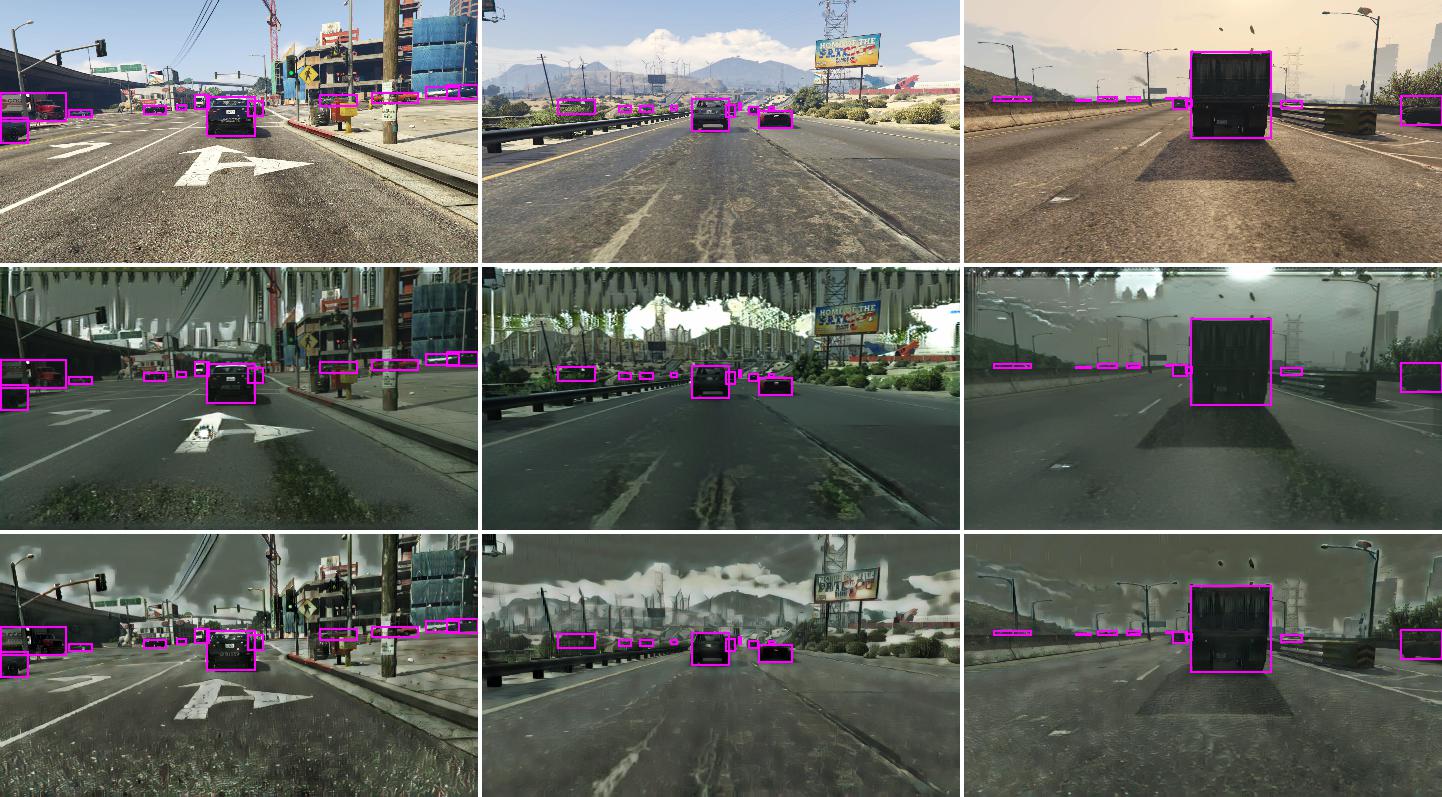}
	\end{subfigure}
	\caption{Samples of translated images with their respective bounding boxes. The real source training images are shown in the first row and their respective fake-target versions are shown in the second and third row for the CycleGAN and AST-AdaIN models, respectively.}
	\label{translation-samples-synthetic}
\end{figure}
Samples of translated images with their respective bounding boxes can be seen in \autoref{translation-samples-synthetic} (more samples available in the supplementary material). As illustrated, the translated models are capable of preserving the global scene structure, as well as the geometry of objects and their co-occurrence, i.e., the geometric relation between the elements from the real to the respective fake image. Despite some unwanted artifacts, as the gray-striped skies observed in the second row, the artificial images proved to be useful for training the detector.

\textbf{Results with VGG-16 and VOC'07.}
\begin{table}[t]
    \centering
    \begin{tabular}{c | c c | c c | c} 
     \toprule
      & \multicolumn{2}{c}{Real} & \multicolumn{2}{|c|}{Fake} & \\
     Method & S & T & C & A & car AP \\
     \midrule\midrule
     Source-only & \cmark & & & & 34.6\\ 
     \hline
     PDA+FDA \citep{shan2019pixel} & \cmark & \ccmark & & & 39.6 \\
     SCL \citep{scl} & \cmark & \ccmark & & & 42.6 \\
     SCDA \citep{scda} & \cmark & \ccmark & & & 43.0 \\
     SAPNet \citep{sapn} & \cmark & \ccmark & & & 44.9 \\
     \hline
     OURS-C & & & \cmark & & 41.9 \\
     OURS-A & & & & \cmark & 45.0 \\
     OURS-C+A & & & \cmark & \cmark & \textbf{46.4}\\
     \hline
     Target-only & & \cmark & & & 61.2 \\
     \bottomrule
    \end{tabular}
    \caption{
    Sim10k$\protect\rightarrow$Cityscapes:
    Results from training with annotated Sim10k (S) and non-annotated Cityscapes (T) using the VGG-16 as backbone, evaluating on Cityscapes with the VOC'07 metric.
    The check-mark denotes which data was used during training, and if crossed, only non-annotated data was used.
    Fake data was acquired by translating Sim10k to Cityscapes using CycleGAN (C) and AST-AdaIN (A).}
    \label{tab:results-synthetic-vgg}
\end{table}
The detection results are shown in \autoref{tab:results-synthetic-vgg}. As it can be seen, our method outperformed source-only and the other models. Although training with the fake-data generated by both translation methods (OURS-C+A) attained the best result with $46.4\%$ in AP, the simpler training settings using the AST-AdaIN method (OURS-A) showed to be sufficient to surpass the other models. This result reinforces the argument that low-level features are more important than high-level ones when adapting from synthetic to real images. Lastly, in this scenario, our method outperformed all state-of-the-art and other known methods for unsupervised domain adaptation for object detection.

\subsubsection{Driving in adverse weather}
In this experiment, the Cityscapes and Foggy Cityscapes datasets are used to evaluate the methods in a similar driving scenario but with a distinct weather condition. Both datasets have the same size, aspect ratio, train/test split, and also have paired instances, only differing by the presence or absence of fog. Although they have paired images, the proposed method does not use this information in order to resemble the real-world scenario where this pairing is virtually impossible.

\textbf{Results.}
\begin{table}[t]
    \centering
    \begin{tabular}{c | c c | c c | c c} 
     \toprule
      & \multicolumn{2}{c}{Real} & \multicolumn{2}{|c}{Fake} & \multicolumn{2}{|c}{mAP}\\
     Method & S & T & C & A & T & S+T\\
     \midrule\midrule
     Source-only & \cmark & & & & 27.1 & 36.3\\ 
     \hline
     DA-Faster \citep{dafaster} & \cmark & \ccmark & & & 33.1 & 37.7\\
     Strong-Weak DA \citep{strongweakda} & \cmark & \ccmark & & & 32.8 & 39.2\\
     \hline
     OURS-C & & & \cmark & & 39.4 & -\\
     OURS-A & & & & \cmark & 30.8 & -\\
     OURS-C+A & & & \cmark & \cmark & \textbf{39.5} & -\\
     OURS-S+C & \cmark & & \cmark & & - & 44.6\\
     OURS-S+A & \cmark & & & \cmark & - & 40.8\\
     OURS-S+C+A & \cmark & & \cmark & \cmark & - & \textbf{44.9} \\
     \hline
     \multirow{2}{*}{Target-only} & & \cmark & & & 44.8 & -\\
     & \cmark & \cmark & & & - & 46.3\\
     \bottomrule
    \end{tabular}
    \caption{
    Cityscapes$\protect\rightarrow$FoggyCityscapes: 
    Results from training with annotated Cityscapes (S) and non-annotated Foggy Cityscapes (T), evaluating solely on Foggy Cityscapes (T) or on both (S+T).
    The check-mark denotes which data was used during training, and if crossed, only non-annotated data was used. Fake data was acquired by translating Cityscapes using CycleGAN (C) and AST-AdaIN (A).}
    \label{tab:results-weather}
\end{table}

\begin{figure}[t!]
	\centering
	\begin{subfigure}{0.75\linewidth}
	    \centering
	    \includegraphics[width=\textwidth]{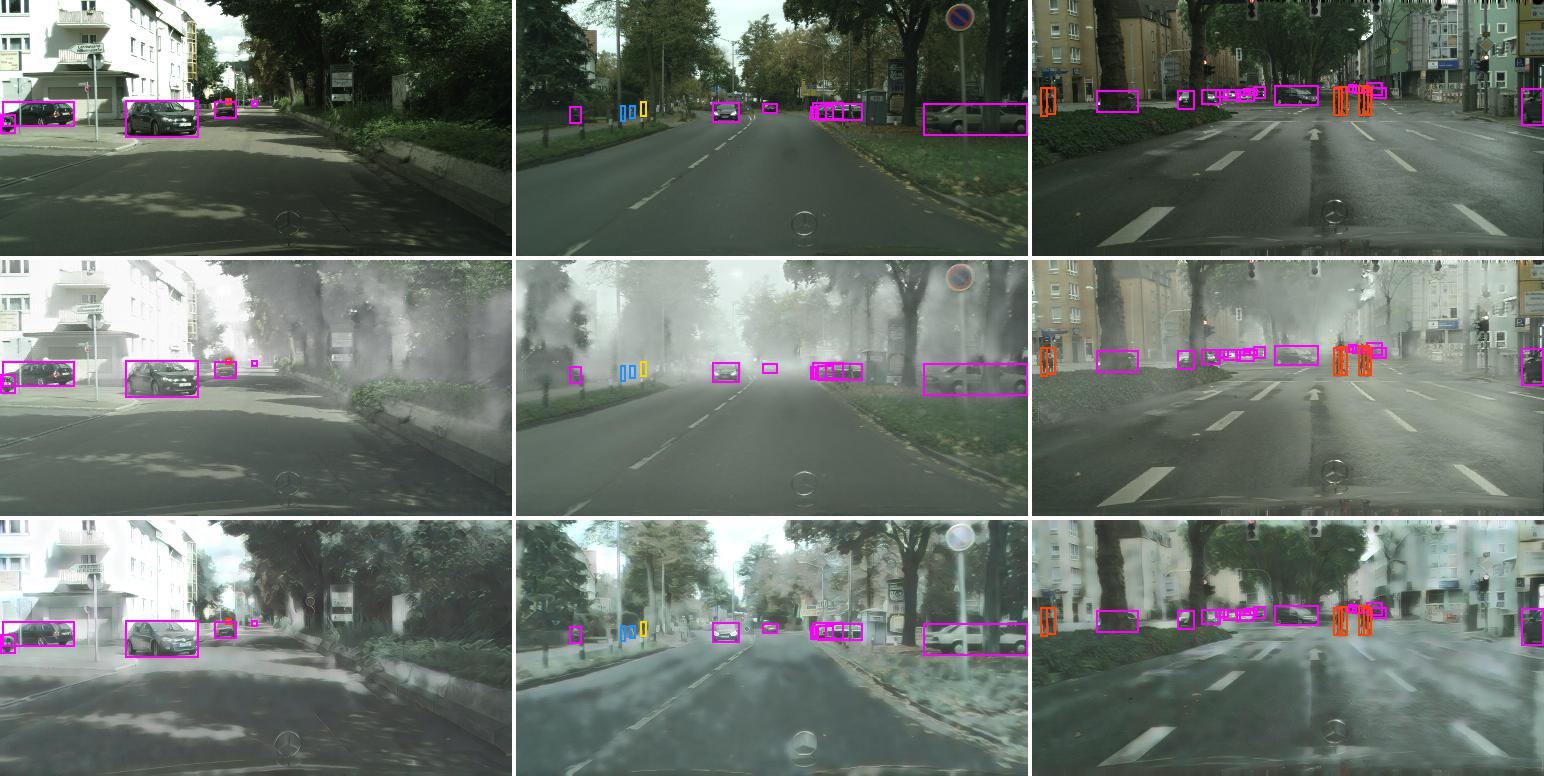}
	\end{subfigure}
	\caption{Samples of translated images with their respective bounding boxes. The real source training images are shown in the first row and their respective fake-target versions are shown in the second and third row for the CycleGAN and AST-AdaIN models, respectively.}
	\label{translation-samples-adverse-weather}
\end{figure}
The results for Cityscapes to Foggy Cityscapes scenario are shown in \autoref{tab:results-weather}. As it can be seen, our method outperformed, again, source-only and both state-of-the-art models. The training setting OURS-C+A surpassed the compared models when evaluated on the target domain, achieving $39.5\%$ in mAP. Evaluating on source and target domains (S+T) simultaneously, representing a scenario in which a single model should handle different weather conditions, our method also outperformed source-only and both the state-of-the-art models. When using real source images together with fake images from both translation models (OURS-S+C+A), our method achieved a performance comparable to the target-only, i.e., below by $1.4$ p.p. ($44.9$ vs. $46.3$). Since the image content (e.g., layout, objects) in this adaptation scenario is the same, a model that strongly translates the domains is expected to perform better, and indeed the CycleGAN-based models achieved better results for generating fake data compared with the AST-AdaIN ones. As it can be seen in \autoref{translation-samples-adverse-weather}, AST-AdaIN struggled to generate realistic foggy images. For the AP of each class, remaining evaluations and more samples of translated images, please refer to our supplementary material.

\subsubsection{Cross-camera adaptation} 
\label{sec:cross-camera}
An object detection system is expected to work well as generally as possible regardless of the hardware setup. In the context of autonomous driving, even datasets collected on similar weather conditions and time of the day may exhibit large domain shifts due to the different image quality and resolution caused by distinct camera settings. Motivated by this fact, this experiment intends to measure how well can a method accurately detect objects on a target dataset whose images have been taken by a different camera setting than the source dataset.

The experiment was carried out using the Cityscapes and KITTI datasets. The datasets were employed in both directions, resulting in two experiments: (i) Cityscapes as source domain and KITTI as target domain; and (ii) KITTI as source domain and Cityscapes as target domain. Taking into account that KITTI has no test set available, two protocols were considered. The first, following the protocol used in \citep{dafaster}, uses KITTI original training partition as the test set for the experiment Cityscapes$\rightarrow$KITTI. However, this protocol does not seem to be fair, given that it uses the training data (without the annotations) as test set. Therefore, for the second protocol, we randomly split the KITTI training set into 70\% and 30\% partitions for the training and test set, resulting in 5237 and 2244 images, respectively.

\begin{table}[t]
    \begin{adjustbox}{width=\columnwidth,center}
    \begin{tabular}{c | c c | c c | c c | c c | c c | c c} 
     \toprule
      & & & & & \multicolumn{4}{c}{car AP (C$\rightarrow$ K)} & \multicolumn{4}{|c}{car AP (K$\rightarrow$ C)}\\
      & \multicolumn{2}{c}{Real} & \multicolumn{2}{|c|}{Fake} & \multicolumn{2}{c|}{Protocol 1} & \multicolumn{2}{c}{Protocol 2} & \multicolumn{2}{|c|}{Protocol 1} & \multicolumn{2}{c}{Protocol 2}\\
     Method & S & T & C & A & T & S+T & T & S+T & T & S+T & T & S+T \\
     \midrule\midrule
     Source-only & \cmark & & & & 74.3 & 71.9 & 74.1 & 68.2 & 29.1 & 80.0 & 28.6 & 65.2\\
     \hline
     DA-Faster \citep{dafaster} & \cmark & \ccmark & & & 71.3 & 68.8 & 70.5 & 64.5 & 31.8 & 81.2 & 30.6 & 66.5\\
     Strong-Weak DA \citep{strongweakda} & \cmark & \ccmark & & & 64.3 & 63.2 & 70.1 & 64.9 & 36.1 & \textbf{82.1} & 35.0 & \textbf{68.8}\\
     \hline
     OURS-C & & & \cmark & & 75.0 & - & 74.8 & - & 34.9 & - & 36.0 & -\\
     OURS-A & & & & \cmark & 72.3 & - & 71.4 & - & \textbf{38.7} & - & \textbf{36.8} & -\\
     OURS-C+A & & & \cmark & \cmark & \textbf{75.9} & - & \textbf{76.5} & - & 34.5 & - & 35.6 & -\\
     OURS-S+C & \cmark & & \cmark & & - & 73.3 & - & 69.4 & - & 81.3 & - & 68.5\\
     OURS-S+A & \cmark & & & \cmark & - & 72.9 & - & 68.8 & - & 80.6 & - & 68.0\\
     OURS-S+C+A & \cmark & & \cmark & \cmark & - & \textbf{73.7} & - & \textbf{70.1} & - & 80.8 & - & 68.3\\
     \hline
     \multirow{2}{*}{Target-only} & & \cmark & & & 88.2 & - & 84.9 & - & 58.9 & - & 58.9 & -\\
      & \cmark & \cmark & & & - & 86.1 & - & 78.3 & - & 86.1 & - & 78.3\\
     \bottomrule
    \end{tabular}
    \end{adjustbox}
    \caption{
    Cityscapes$\protect\rightarrow$KITTI (C$\protect\rightarrow$K): 
    Results from training with annotated Cityscapes (S) and non-annotated KITTI (T), evaluating solely on KITTI (T) or on both (S+T).
    KITTI$\protect\rightarrow$Cityscapes (K$\protect\rightarrow$C): 
    Results from training with annotated KITTI (S) and non-annotated Cityscapes (T), evaluating solely on Cityscapes (T) or on both (S+T).
    The check-mark denotes which data was used during training, and if crossed, only non-annotated data was used. Fake data was acquired by translating the source training set using CycleGAN (C) and AST-AdaIN (A).}
    \label{tab:results-c-k}
\end{table}
\begin{figure}[t!]
	\centering
	\begin{subfigure}{0.75\linewidth}
	    \centering
	    \includegraphics[width=\textwidth]{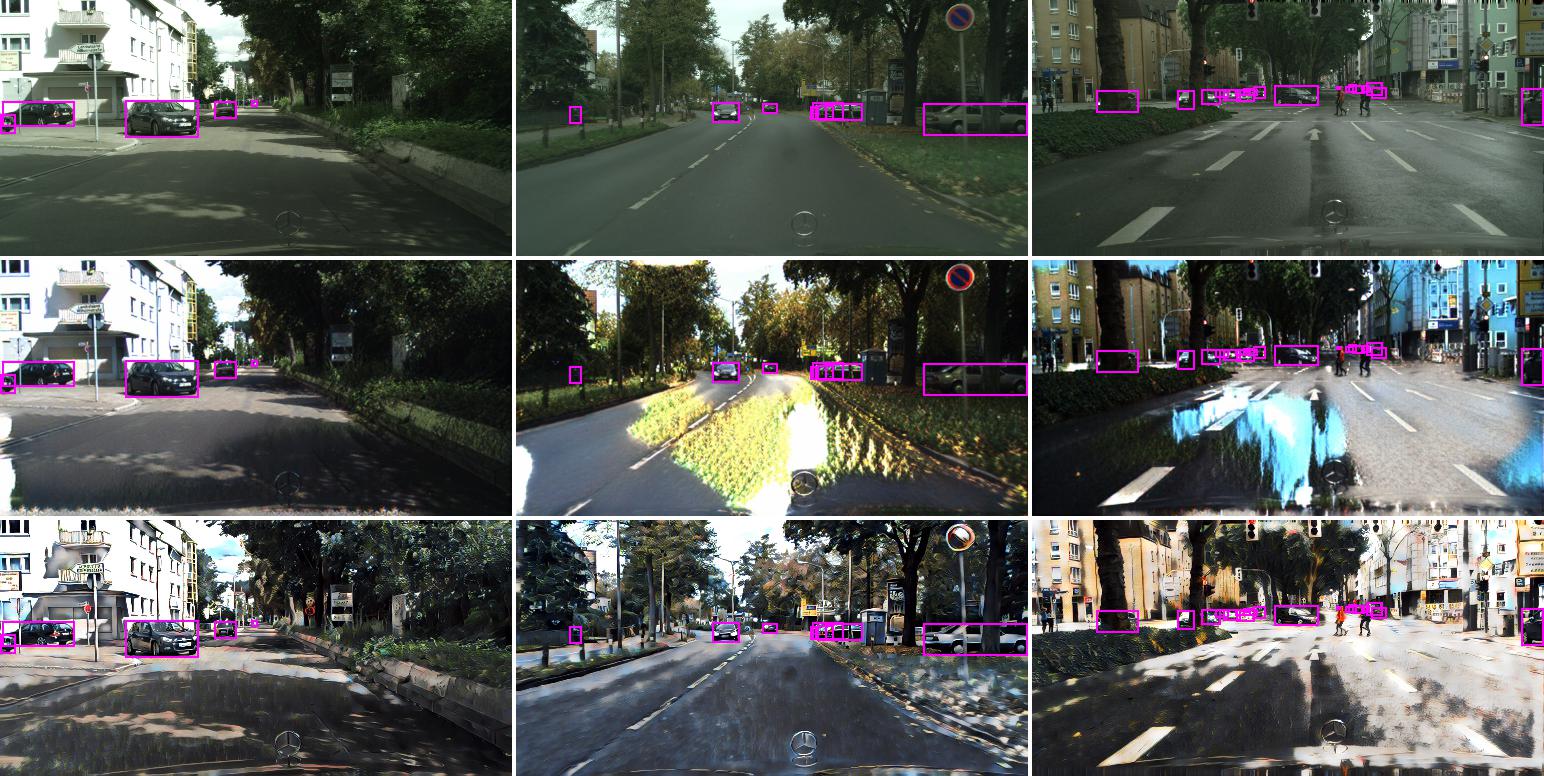}
	    \includegraphics[width=\textwidth]{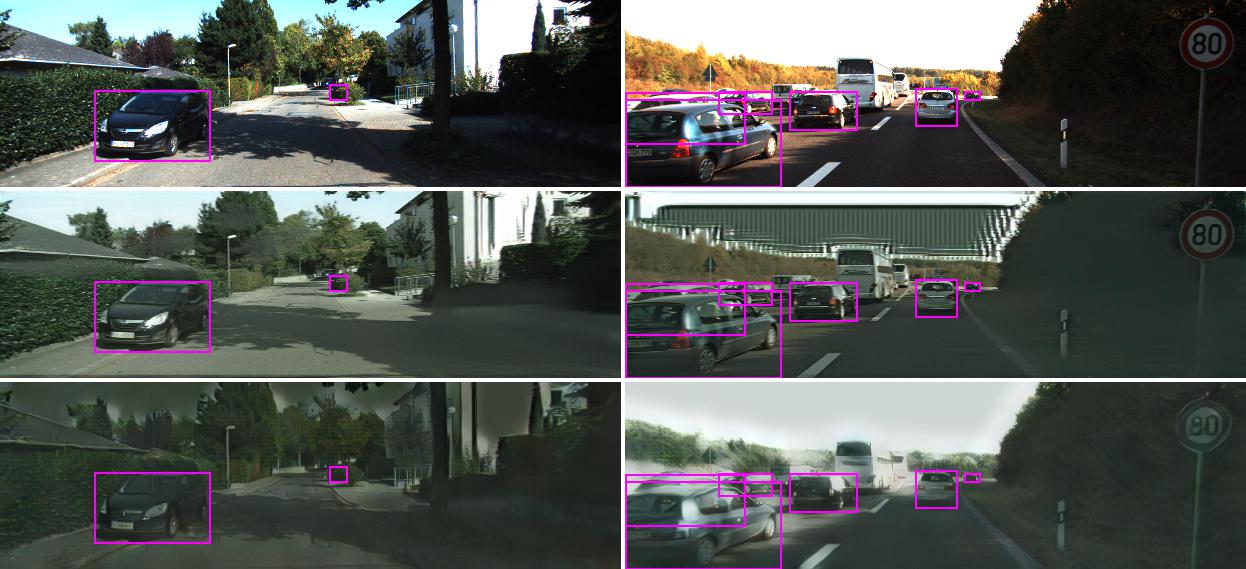}
	\end{subfigure}
	\caption{Samples of translated images with their respective bounding boxes. Top: the translation from Cityscapes to KITTI, and vice-versa for the bottom one. The real source training images are shown in the first row and their respective fake-target versions are shown in the second and third row for the CycleGAN and AST-AdaIN models, respectively.}
	\label{translation-samples-camera}
\end{figure}

\textbf{Results (Cityscapes$\rightarrow$KITTI).} As shown in \autoref{tab:results-c-k}, our method outperformed source-only and both state-of-the-art models in both protocols. Overall, although the translation produced unwanted artifacts (\autoref{translation-samples-camera}, more samples available in the supplementary material) such as the stain over the asphalt, the CycleGAN yielded better results than AST-AdaIN. This may be because KITTI has richer colors and textures in contrast to Cityscapes (see \autoref{fig:dataset_samples}), requiring a more complex model to translate the image domain. Nevertheless, although the results of OURS-A were below source-only (suggesting that the AST-AdaIN model did not transfer the style properly), when jointly trained with fake data from CycleGAN (OURS-C+A), it added significant information to the training, improving the AP from $75.0\%$ to $75.9\%$ and $74.8\%$ to $76.5\%$ when testing on target with protocol 1 and 2, respectively.

\textbf{Results (KITTI$\rightarrow$Cityscapes).}
In the opposite direction of the previous scenario, our method outperformed source-only and both state-of-the-art models when testing on target domain. In contrast to previous scenario (Cityscapes$\rightarrow$KITTI), the setting OURS-A produced (\autoref{tab:results-c-k}) the best results on both protocols when evaluating on the target domain, being suitable when targeting a less complex dataset, such as Cityscapes where images resembles more opaque colors when compared to KITTI.

Among the compared models, Strong-Weak-DA achieved the highest mAP, with $2.6$ and $1.8$ p.p. below our best results on protocol 1 and 2 ($36.1$ vs. $38.7$ and $35.0$ vs. $36.8$), respectively. However, when evaluating on both domains, our best results achieved $81.3$ and $68.5$ compared to
$82.1$ and $68.8$ from Strong-Weak-DA for protocols 1 and 2, respectively; a difference of only $0.8$ and $0.3$ p.p.. Although our method did not achieve the highest mAP in this last experiment, the small difference still makes our method comparable. Regarding the protocols, protocol 1 reflects the results of the state-of-the-art methods, however, protocol 2 depicts a fairer setup by partitioning the data into train and test properly. Still, our results on both protocols remain consistent throughout the training settings. The remaining mAP results are available in the supplementary material.

\subsection{Limitations}
The effectiveness of the proposed method seems to be related to the quality of the translated images, i.e., how much the translated images resemble the target domain. Therefore, the limitations of the methods used for the image translation also apply to our approach. For instance, it is well-known that GAN-based models present unstable and hard-to-converge training, which may result in mode collapse, requiring re-training. Moreover, training these models is time-consuming, mostly because of their substantial amount of parameters, as in CycleGAN. In contrast, style-transfer methods not based on GANs, such as AST-AdaIN, are stable and less time-consuming. However, they are limited to translating only colors and textures without concerning about the semantics of the scene, being effective only for small domain-shifts. Moreover, they do not allow verifying the quality of the adaption as in two stage methods.

In addition, the proposed method may be limited to object sizes seen by the detector during training. If objects in the source domain have different sizes in the target domain, the effectiveness of the method will depend on the capability of the chosen object detector to work on object sizes not seen during training.

In this work, we did not include all results for the methods proposed in the works \citep{shan2019pixel, scl, sapn, scda} due to the unavailability of the source-code or because it requires editing it to match our settings. Instead, we adapted our proposed method to match their settings (Section 4.5.1).

\section{Conclusion}
\label{sec:conclusion}
This paper addressed the detection of objects across distinct domains by training a detector on annotated data from a source domain and non-annotated data from a target domain, which poses an Unsupervised Domain Adaptation problem. The proposed method tackles this challenging problem using a simpler approach compared to the state-of-the-art, yet it achieves comparable or, in most cases, higher performance in several scenarios. The method uses unsupervised image-to-image translation and style-transfer models to generate annotated data in the target domain that are later used to train a deep object detector. The proposed two-stage approach has the benefit of giving access to the translated training images, from the source to the target domain at pixel-level. Therefore, it allows a visual inspection of the images before training the detector. Moreover, as the two stages are decoupled, our approach does not require additional GPU memory compared to the original object detector. In contrast, the state-of-the-art methods are single-stage approaches that aimed at minimizing the domain discrepancy at the cost of adding more complexity into their models and less interpretability.

The evaluation was performed on well-known relevant datasets for object detection in autonomous driving that resembles a diverse and realistic set of scenarios, including learning from synthetic data, driving in adverse weather, and cross-camera adaptation. In addition, we carried out an extensive comparison with the state-of-the-art. Following, we provide a summary of the results aggregating our main achievements. Our method outperformed the state-of-the-art models by up to $9.9$ p.p. in mAP. When learning from synthetic data, we outperformed the source-only and both state-of-the-art models on all training settings (for this experiment: OURS-C, OURS-A and OURS-C+A) from $7.7$ to $9.9$ p.p. in mAP improvement. For driving in adverse weather, we achieved $1.6$ to $6.4$ p.p. superior in five out of six different training settings. On the cross-camera adaptation, considering the two protocols, our method outperformed the source-only and both state-of-the-art models in 10 out of 12 training settings for the Cityscape to KITTI scenario, and 4 out of 12 training settings for the KITTI to Cityscapes scenario. An increase ranging from $0.6$ to $2.6$ p.p. in mAP.

The results show that our method is simple yet effective, outperforming the current state-of-the-art models in all three scenarios, reducing the gap toward the upper-bound. Moreover, the results demonstrate that our method can benefit from the fake-data generation models even when the qualitative results seem inaccurate and show some unwanted artifacts.

There is still room for improvements. For instance, in future studies, one should investigate alternative methods for image translation and style transfer, especially those that could improve the semantic consistency of the translation. This will likely boost the performance of the object detector. In addition, future works should collect and experiment with new datasets, preferably in other domains and scenarios less related to the driving environment.

\section*{Conflict of interest}
The authors declare that they have no conflict of interest.

\section*{Acknowledgements}
This study was financed in part by Coordenação de Aperfeiçoamento de Pessoal de Nível Superior (CAPES, Brazil) - Finance Code 001; Conselho Nacional de Desenvolvimento Científico e Tecnológico (CNPq, Brazil) - grants 311654/2019-3, 311504/2017-5 and 200864/2019-0; and Fundação de Amparo à Pesquisa do Espírito Santo (FAPES, Brazil) - grant 84412844. We thank the NVIDIA Corporation for the donation of a Titan Xp GPU used in this research.

\bibliography{refs}

\end{document}